\documentclass[preprint,authoryear,12pt]{elsarticle}
\usepackage{times}
\usepackage{graphicx}
\usepackage{wrapfig}
\usepackage{sidecap}
\usepackage{amsmath,amssymb} 
\usepackage{afterpage}
\usepackage[]{subfigure}
\usepackage{mdwtab}
\usepackage{caption}
\usepackage{color}
\usepackage{times}
\usepackage{algorithm}
\usepackage{algorithmic}
\usepackage{mathtools}
\DeclarePairedDelimiter\ceil{\lceil}{\rceil}

\graphicspath{{figures/}}

\usepackage{etex}
\reserveinserts{18}
\usepackage{morefloats}

\usepackage{amssymb}

\journal{Computerized Medical Imaging and Graphics}

\begin{document}

\begin{frontmatter}

\title{Automatic tracking of vessel-like structures from a single starting point}

\author[label1,label2]{D\'ario Augusto Borges Oliveira\corref{cor1}}
\address[label1]{Institute of Mathematics and Statistics, University of S{\~a}o Paulo}
\address[label2]{Electrical Engineering Department, Pontifical Catholic University of Rio de Janeiro}

\cortext[cor1]{I am corresponding author}

\ead{darioaugusto@gmail.com}

\author[label3]{Laura Leal-Taix\'e}
\address[label3]{Institute for Information Processing, Leibniz University Hannover}

\author[label2,label4]{Raul Queiroz Feitosa}
\address[label4]{Electrical Engineering Department, Rio de Janeiro State University}

\author[label3]{Bodo Rosenhahn}

\begin{abstract}

The identification of vascular networks is an important topic in the medical image analysis community. While most methods focus on single vessel tracking, the few solutions that exist for tracking complete vascular networks are usually computationally intensive and require a lot of user interaction. In this paper we present a method to track full vascular networks iteratively using a single starting point. Our approach is based on a cloud of sampling points distributed over concentric spherical layers. We also proposed a vessel model and a metric of how well a sample point fits this model. Then, we implement the network tracking as a min-cost flow problem, and propose a novel optimization scheme to iteratively track the vessel structure by inherently handling bifurcations and paths. The method was tested using both synthetic and real images. On the 9 different data-sets of synthetic blood vessels, we achieved maximum accuracies of more than 98\%. We further use the synthetic data-set to analyse the sensibility of our method to parameter setting, showing the robustness of the proposed algorithm. For real images, we used coronary, carotid and pulmonary data to segment vascular structures and present the visual results. Still for real images, we present numerical and visual results for networks of nerve fibers in the olfactory system. Further visual results also show the potential of our approach for identifying vascular networks topologies. The presented method delivers good results for the several different datasets tested and have potential for segmenting vessel-like structures. Also, the topology information, inherently extracted, can be used for further analysis to computed aided diagnosis and surgical planning. Finally, the method's modular aspect holds potential for problem-oriented adjustments and improvements.

\end{abstract}

\begin{keyword}
vascular network tracking \sep linear programming \sep vessel characterization \sep medical imaging
\end{keyword}

\end{frontmatter}



\section{Introduction}

The identification of vascular networks is a topic of general interest in medical image analysis and
in particular for diagnosis of problems related to the vascular system such as cerebrovascular
accidents or thrombosis. While efficient algorithms for single vessel tracking exist,
the segmentation of complete vascular networks is still a challenge, mainly due to huge search
space involved. Indeed, most solutions are only locally optimal, very computationally intensive or
both.

We refer the interested reader to \cite{lesagemia2009} for a detailed review of methods to tackle the vascular segmentation problem. Many of these works are based on a propagating
structure emanating from a given starting point using different techniques such as level-sets
\cite{lorigo2001,luboz2005} or minimal paths \cite{deschamps2001,wink2000}. Other approaches are based on particle filters \cite{florin2005,lesage2008}, Markov chain processes \cite{lacoste2006}, statistical methods for tubular structures \cite{wang2012} and multiple hypothesis testing \cite{frimanmia2010}. Recent works \cite{turetkenneuro2011,turetkencvpr2012} have shown that global optimization can be
reliable as it finds all the vessels of the structure jointly. In \cite{turetkenneuro2011}, authors find a set of vessel candidate points and then solve the $k$-minimum spanning tree ($k$-MST) problem to find the final structure, while in \cite{turetkencvpr2012}, the vessel detection model needs to be specifically trained for each type of data. The latter assumes a well behaved distribution of vessel points and a careful training. 


In this paper we present a novel Linear Programming (LP) solution for tracking vascular
networks through a fast iterative method that tracks each vessel branch independently and
is able to handle bifurcations. Our approach does not impose any restrictions to the form of
vessels’ offshoots. It relies on  two simple assumptions: vessels are nearly cylindrical, having circular or elliptic cross sections, and present a tree like structure. Contrary to most methods proposed thus far, it only requires
one starting point to track a full vessel network with arbitrary form.

\subsection{Motivation and Graph Modeling}
\label{sec:method:motivation}

\begin{figure}[t]
\centering
\includegraphics[scale=0.5]{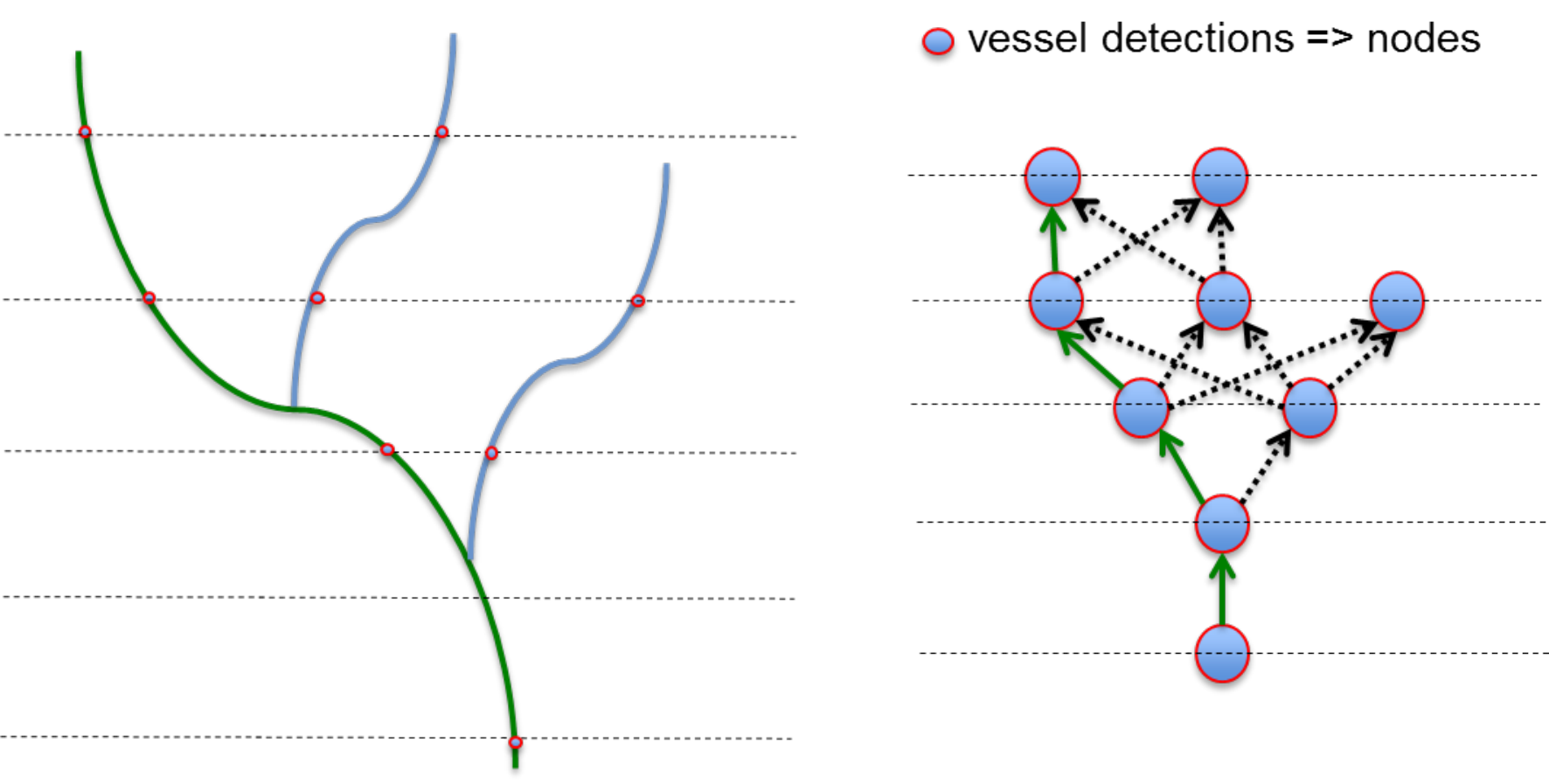}
\caption{Modeling a vascular network in a graph using a single direction. On the left side there is a model representing a vascular network with an identified green branch; on the right side its respective graph with green edges indicating the green branch path, and dashed edges representing other path possibilites.}
\label{fig:fig1}
\end{figure}

The use of directed graphs to segment tree-like structures (such as vascular networks) is somewhat intuitive. Considering vessel point detections as graph nodes, a vascular network can be structured as a directed graph, which allows the segmentation of its branches through the analysis of graph paths, as depicted in Figure \ref{fig:fig1}.

\begin{figure}[t]
\centering
\includegraphics[scale=0.5]{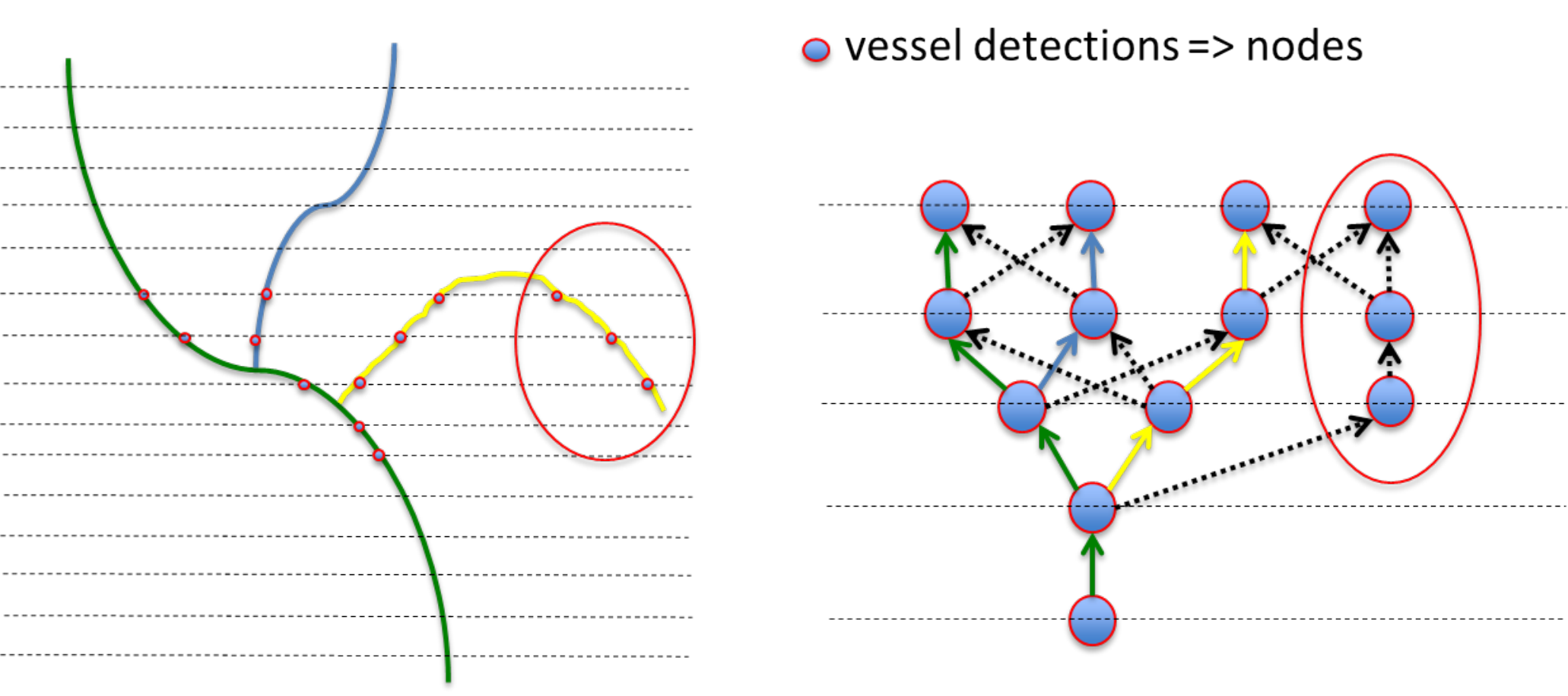}
\caption{Issues of modeling a vascular network as a graph using a single direction (some edges were hidden to improve understanding). The model on the left side has its green and blue branches correctly modelled in the graph represented on the right side. The yellow path, though, is incorrectly represented - as the red line points out - due to directed graph representation issues.}
\label{fig:fig2}
\end{figure}

Not very intuitive is the problem of using directed graphs to represent vascular structures in images with voxels distributed uniformly in a given direction, such as CT scans. Vessels usually change direction continuously and hardly follow the direction of a single axis. Since directed graphs consist of nodes organized in interconnected layers following a certain direction, this behaviour is problematic. The use of a single axis to define the directed graph levels, would lead to the impossibility of segmenting vessels whose direction goes in favor of the axis at some point, but against it later. In other words, the directed graph modeling does not allow a path passing through a sequence of levels N and N+1 to return to any node on level N. Figure \ref{fig:fig2} depicts the problem. The yellow branch cannot be entirely represented since it crosses some planes more than once.

\begin{figure}[t]
\centering
\includegraphics[scale=0.5]{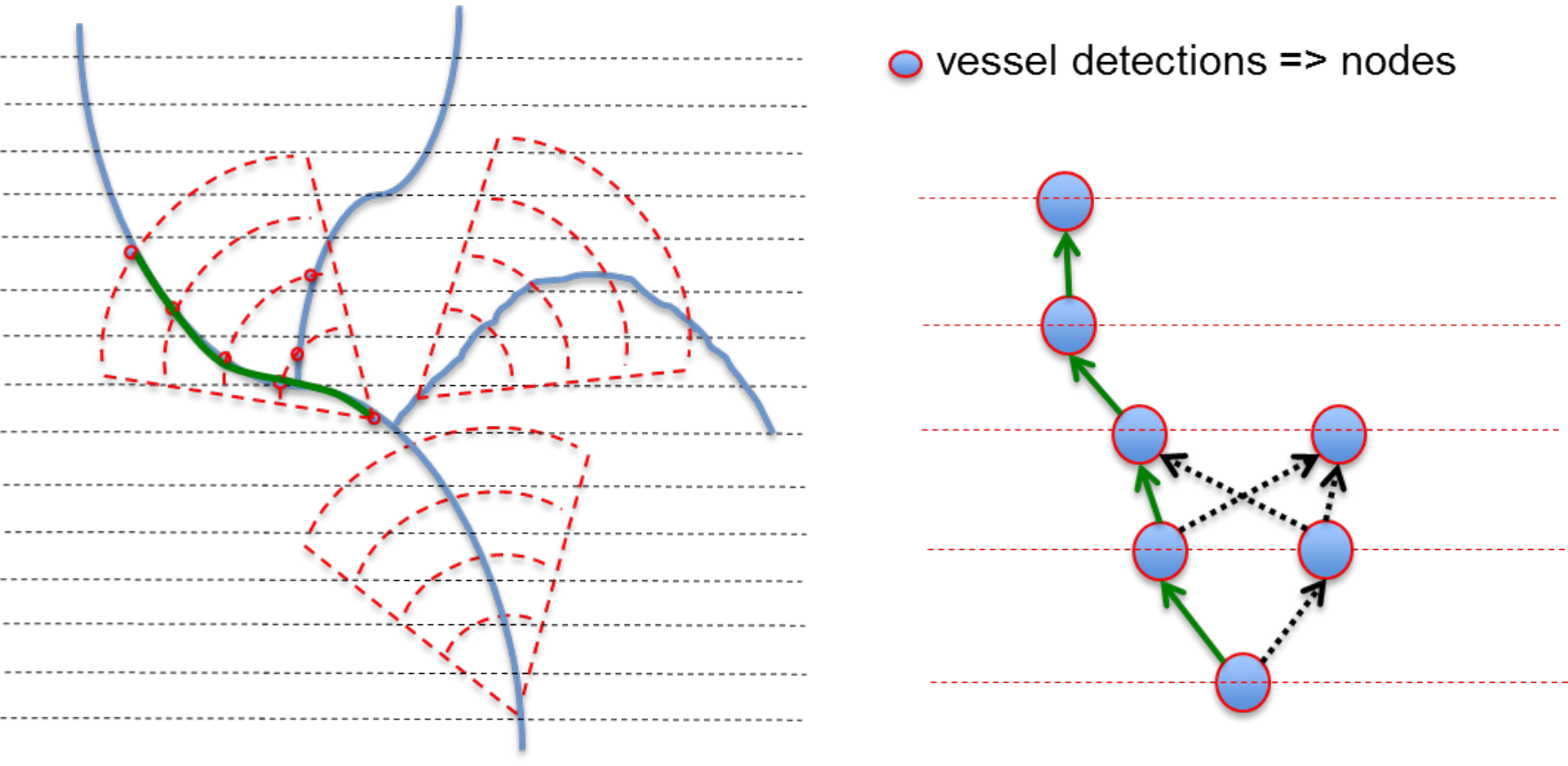}
\caption{Definition of graphs using the local vessel direction of the vascular network. The use of vessel-driven sampling models allows the corrected use of directed graph models, overcoming the representation issues showed in Figure 2.}
\label{fig:fig3}
\end{figure}

This paper proposes a way to overcome this difficulty  through the use of multiple graphs, each one modelling a portion of the vascular network that conforms the implied sequential spatial arrangement of vessel detection. These
local graphs are created iteratively and their nodes are associated to spatial  positions expressed in a convenient coordinate system, as illustrated in figure \ref{fig:fig3}. This approach provides, at least locally, a model in which the problem described in figure \ref{fig:fig2} does not occur, so that the modelling through directed graphs becomes
effective. The proposed sampling model is formally defined in section \ref{sec:method:sampling}.

Our contribution is three-fold:

\begin{itemize}
\renewcommand{\labelitemi}{$\bullet$}
\item{Oriented conical sampling method using vessel point detections to allow tracking of each vessel along its axis.}
\item{New formulation of the vascular network tracking problem as a min-cost flow problem to track local vascular structures from a single seed point while dealing with bifurcations.}
\item{Novel optimization scheme that iteratively tracks the full vascular network and guarantee anatomical vascular properties.}
\end{itemize}

Our paper is structured as the following: in section \ref{sec:method} we explain in detail our method, in section \ref{sec:resultado:resultados} we show and discuss the results, and in section \ref{sec:conclusions} we present our conclusions.

\afterpage{\clearpage}

\section{Materials and Methods}
\label{sec:method}

\subsection{General Algorithm Description}
\label{sec:method:algorithm}

\begin{figure}[t]
\centering
\includegraphics[scale=0.4]{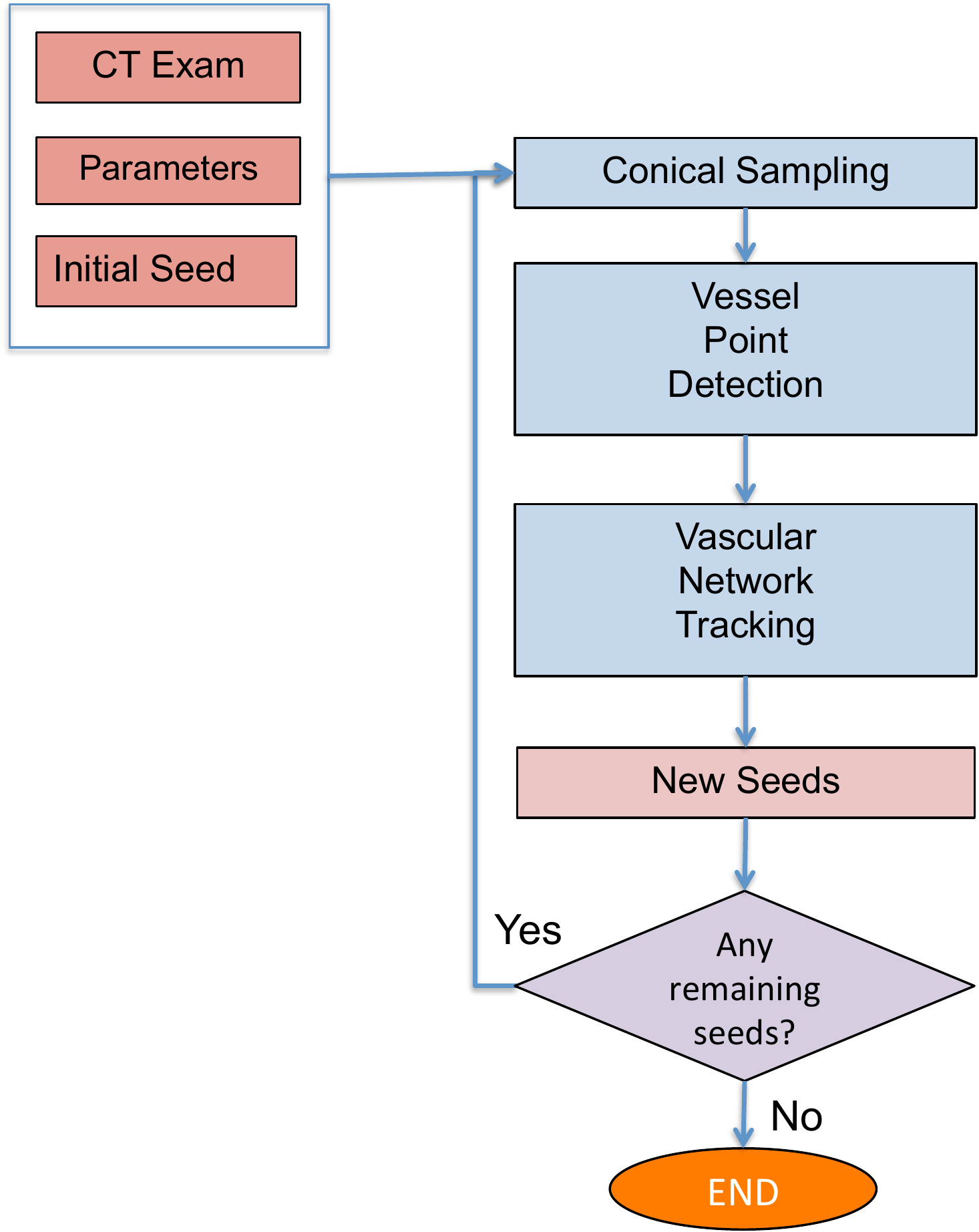}
\caption{Flowchart of the proposed method.}
\label{fig:flowchart}
\end{figure}

The proposed method starts at a single user-defined vessel point, called initial seed. From this
seed, emerging branches of the vascular network are detected, each one deriving a new seed
and possibly new branches. Subsequent iterations are executed until the full vascular
network is segmented, as outlined in figure \ref{fig:flowchart}.

The full procedure involves the following steps:
\begin{enumerate}
	\item Initial seed point definition. In this step a initial seed is defined to be used at the starting point for the whole vascular network to be found. The definition of this point can be done manually by the user or using any automatic procedure.
	\item Sampling model creation. Here a cloud of sample points is created within a conical neighborhood with apex at a given seed point.
	\item Vessel point detection. In this step we test all sample points within the cloud to select those very likely to belong to a vessel.
	\item Vascular network identification. Here we use the vessel point candidates selected in the previous step to track a local vascular network using a proposed network flow analysis approach.
	\item New seeds definition. In this step new seed points are created from the vascular network branches found in the previous step. 
	\item Repeat steps 2 to 5 until there is no new seed to be evaluated.
\end{enumerate}

Each of the steps from step 2 on is explained in detail in the following sections.

\subsection{Sampling Model Creation}
\label{sec:method:sampling}

The sampling method is important for the procedure outlined in previous section, as it determines the accuracy as well as the computational efficiency of our proposal. In this section we present our procedure for this matter. It is important to mention that any other sampling procedures that fulfill the requirements of generating a nearly uniform distribution of sample points over a conical neighborhood emerging from a seed point, could be used.

\begin{figure}[t]
\centering
\includegraphics[scale=0.5]{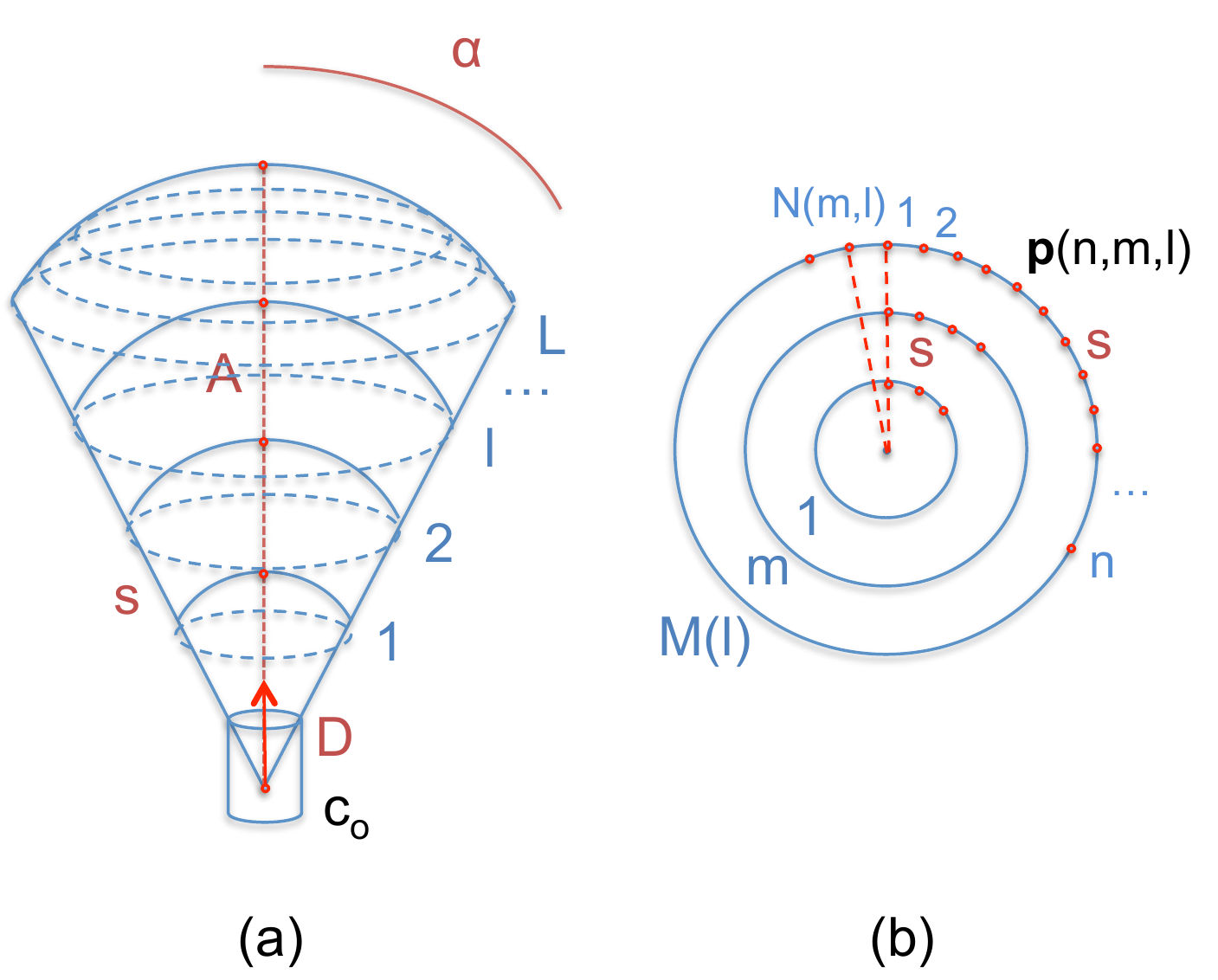}
\caption{Conical sampling model structured by means of a set of spherical calottes.}
\label{fig:conical}
\end{figure}

Figure \ref{fig:conical} describes the sampling model. The cylinder at the bottom of Fig. \ref{fig:conical}a represents a cylindrical elementary vessel section centered at the current seed, which is taken as origin of a coordinate system having $(x,y,z)$ as its orthonormal basis, whereby the unitary vector $z$ is aligned with the vessel central axis.  

A conical neighborhood with apex in the origin (see fig. \ref{fig:conical}a) will be explored for vessel points. It is defined by two parameters, namely:
\begin{itemize}
\item the aperture angle $\alpha$, and
\item the height $A$.
\end{itemize}

Instead of examining all voxels in the neighborhood, only points over spherical calottes centered at the origin with radii equal to $ls$ are considered (see Figure \ref{fig:conical2}), where $s$ is a user-defined parameter specifying the separation between any adjacent calottes, and  $l$ is an integer number between 1 and $L$, the number of calottes given as a further input parameter. Hence, the height $A$ is determined by equation \ref{eq:A}:

\begin{figure}[t]
\centering
\includegraphics[scale=0.5]{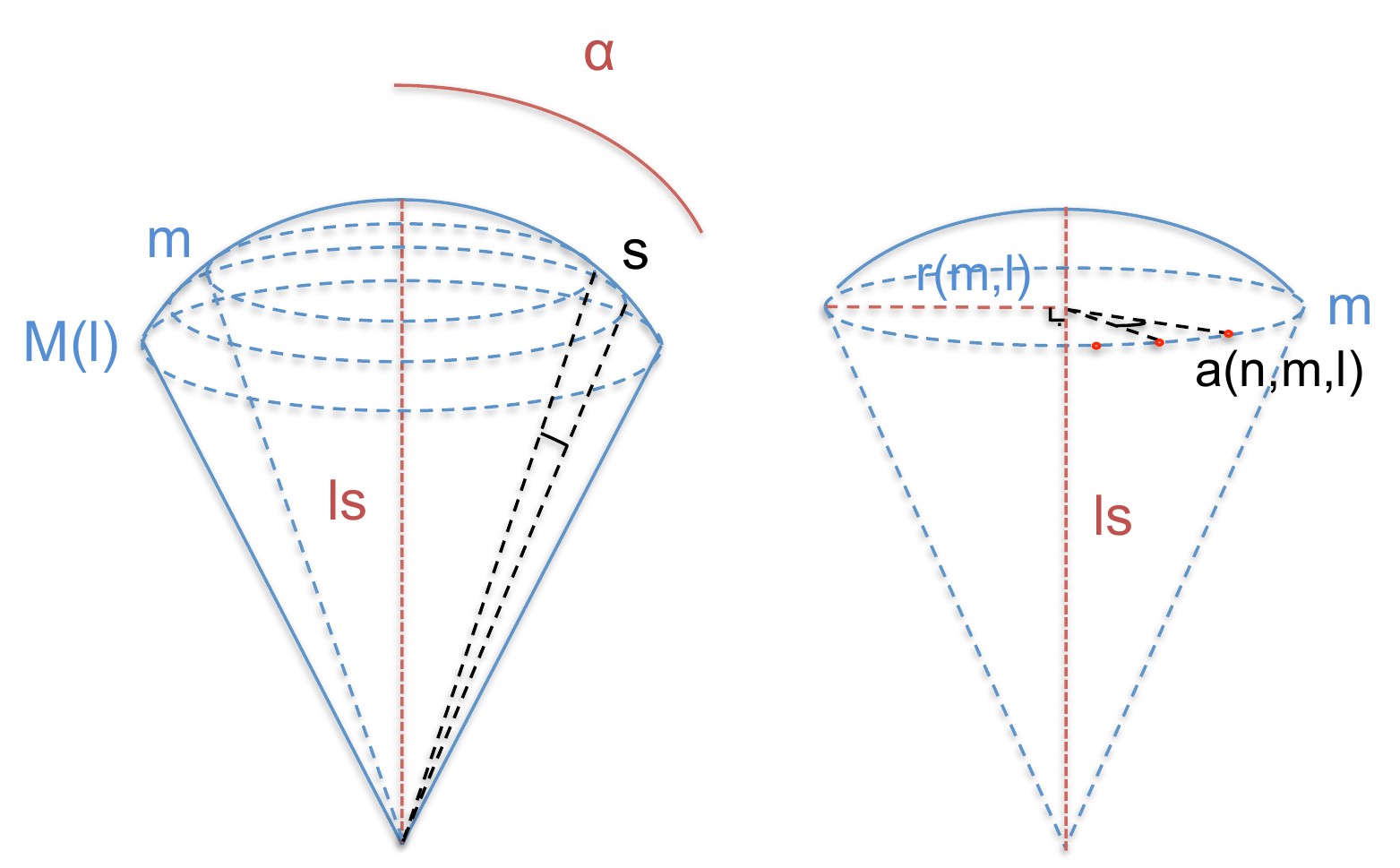}
\caption{The values of $\alpha$ and the sampling distance $s$ define the $M(l)$ sampling circles at a given layer. The radius $r(m,l)$ and the sampling distance $s$ define the number of sample points in each circumference $m$.}
\label{fig:conical2}
\end{figure}

\small
\begin{align}
\label{eq:A}
A = Ls
\end{align}

\normalsize

Over each calotte $l$, $M(l)$ circumferences are determined as shown in Figure \ref{fig:conical}b and defined in equation \ref{eq:M}, so that the minimum distance between any two points lying on consecutive circumferences over the same calotte is at most equal to $s$ (actually a value close to it). 

\small
\begin{align}
\label{eq:M}
\textstyle
M(l) = \ceil*{l\alpha} + 1
\end{align}
\normalsize

Notice that the number of circumferences determined over a calotte increases with  the calotte radius. The radius $r(m,l)$ of the $m^{th}$ circumference over the $l^{th}$ calotte (see Figure \ref{fig:conical2}) is given by equation \ref{eq:rml}:

\small
\begin{align}
\label{eq:rml}
\textstyle
r(m,l) = ls\cos(\frac{\pi}{2}-\frac{m}{l}) && \textrm{for} \ l \in \{1..L\} \ \textrm{and} \ m \in \{1..(M(l)-1)\} 
\end{align}
\normalsize


Notice that for $m=0$, the circumference boils down to a single point lying on the cone central axis.

Sample points are uniformly distributed along each circumference, whereby the arc determined by two adjacent sample points is at most equal to $s$. Thus, the number $N(m,l)$ of sample points uniformly distributed along the $m^{th}$ circumference of the $l^{th}$ calotte is given by equation \ref{eq:nml}:

\small
\begin{align}
\label{eq:nml}
\begin{split}
\textstyle
N(m,l) = \ceil*{\frac{2\pi r(m,l)}{s}} = \ceil*{2\pi l\cos(\frac{\pi}{2}-\frac{m}{l})} \ \textrm{for} \ l \in \{1..L\} \ \textrm{and} \ m \in \{1..(M(l)-1)\} 
\end{split}
\end{align}
\normalsize

Hence, the actual arc length $a(m,l)$ determined by adjacent sample points along the $m^{th}$ circumference of the $l^{th}$ calotte will be given by equation \ref{eq:aml}:

\small
\begin{align}
\label{eq:aml}
\textstyle
a(m,l) = \frac{2\pi}{ml} && \textrm{for} \ l \in \{1..L\} \ \textrm{and} \ m \in \{1..(M(l)-1)\} 
\end{align}
\normalsize

Recall that the unitary vector $\bold{z}$ is aligned with the conic central axis, and $\bold{x,y,z}$ form an orthonormal basis. It can be easily demonstrated that the coordinate vector $\bold{p}(n,m,l)$ of the $n^{th}$ sample point over the $m^{th}$ circumference of the $l^{th}$ calotte can be given by equation \ref{eq:pnml}:

\small
\begin{align}
\label{eq:pnml}
\bold{p}(n,m,l) = \begin{bmatrix} x(n,m,l) \\  y(n,m,l) \\  z(n,m,l) \end{bmatrix} = \begin{bmatrix} \cos{(n\frac{a(m,l)}{r(m,l)})} \cos{(ls)} \\  \sin{(n\frac{a(m,l)}{r(m,l)})}\cos{(ls)} \\  \cos{(ls)} \end{bmatrix} 
\end{align}
\normalsize

where $r(m,l)$ and $a(m,l)$ are given by equations \ref{eq:rml} and \ref{eq:aml}, respectively.

Once the basic spatial distribution of sample points has been defined, we need a geometric transformation that places the point cloud properly on the neighborhood of the seed being considered. Basically, the conic points cloud must be rotated so that the cone central axis, aligns with vessel's central axis at the seed point position, and its origin shifted to the seed point position.


This is done by first multiplying the coordinates given in equation \ref{eq:pnml} by a 3x3 rotation matrix $\mathcal{R}(\textbf{D})$ having as its third row the transpose of the unitary vector $\textbf{D}=[Dx, Dy, Dz]^\intercal$ aligned with the vessel central axis at the seed, and then, by adding the result to the seed coordinate vector. The first and second rows of $\mathcal{R}(\textbf{D})$ are arbitrary, provided that $\mathcal{R}(\textbf{D})$ is unitary. Hence, we define $\mathcal{R}(\textbf{D})$ as:

\small
\begin{align}
\label{eq:rotationmatrix}
\mathcal{R}(\textbf{D}) = \begin{bmatrix} \frac{-D_y}{|V_1|} & \frac{D_x}{|V_1|} & 0 \\  \frac{-D_z D_x}{|V_2|}  & \frac{-D_z D_y}{|V_2|} & \frac{D_x D_x + D_y D_y}{|V_2|} \\  D_x & D_y & D_z \end{bmatrix}
\end{align}
\normalsize

where $V_1$ is an arbitrary unitary vector orthogonal to $\textbf{D}$ and $V_2 = \textbf{D} \times V_1$.

This procedure creates a canonical cloud of points, at each seed point, a cloud of points in this shape is created by translating and rotating this model accordingly fitting the seed point position and direction.

Finally, the coordinate vector $\bold{p}^\star$ of sample points associated to a seed located at $\bold{c}$, whose vessel section is oriented in the direction defined by the unitary vector $\textbf{D}$ is given by:

\small
\begin{align}
\label{eq:p-star}
\textstyle
\bold{p}^\star (n,m,l,D,\bold{c}) = \mathcal{R}(\textbf{D}) \bold{p}(n,m,l) + \bold{c}
\end{align}
\normalsize

where $\mathcal{R}(\textbf{D})$ is given by equation \ref{eq:rotationmatrix} and $\bold{p}(n,m,l)$ is defined in equation \ref{eq:pnml}.

The spatial distribution of sample points in the proposed model (figure \ref{fig:conical}) presents some interesting characteristics:

\begin{itemize}
    \item The probability that a voxel belongs to the vessel network diminishes for layers with larger radii. 
    \item The area of a layer grows with its distance to the seed. This complies with the expected increase of vessel network spreading as one moves away from the seed.
    \item The cone opening angle and height are related to the potential of vessel to change direction suddenly.
\end{itemize}

Once the sample point coordinates have been defined, each of them is evaluated as a vessel point candidate, using the vesselness metric proposed in the next section. 




\subsection{Vessel Point Detection}
\label{sec:method:detection}

This section describes the procedure to assess how well sample points qualify as vessels. 
It involves two sequential steps: (a) computation of a metric, hereafter called vesselness,
to assess how well a sample point qualifies as  part of a vessel; (b) selection of points based on local characteristics, including vesselness. These steps
are described in detail in the following subsections.

\subsubsection{Vesselness Computation}
\label{sec:method:detection:computation}

Recalling section \ref{sec:method:motivation}, sample points are associated to vessel hypotheses according to a determinate vessel model. Different vessel models can be found in the literature, such as elliptical cross-sections models \cite{florin2005}, spheres \cite{rossignac2007} and template models \cite{frimanmia2010,worz2007}.

\begin{figure}[t]
\centering
\includegraphics[scale=0.5]{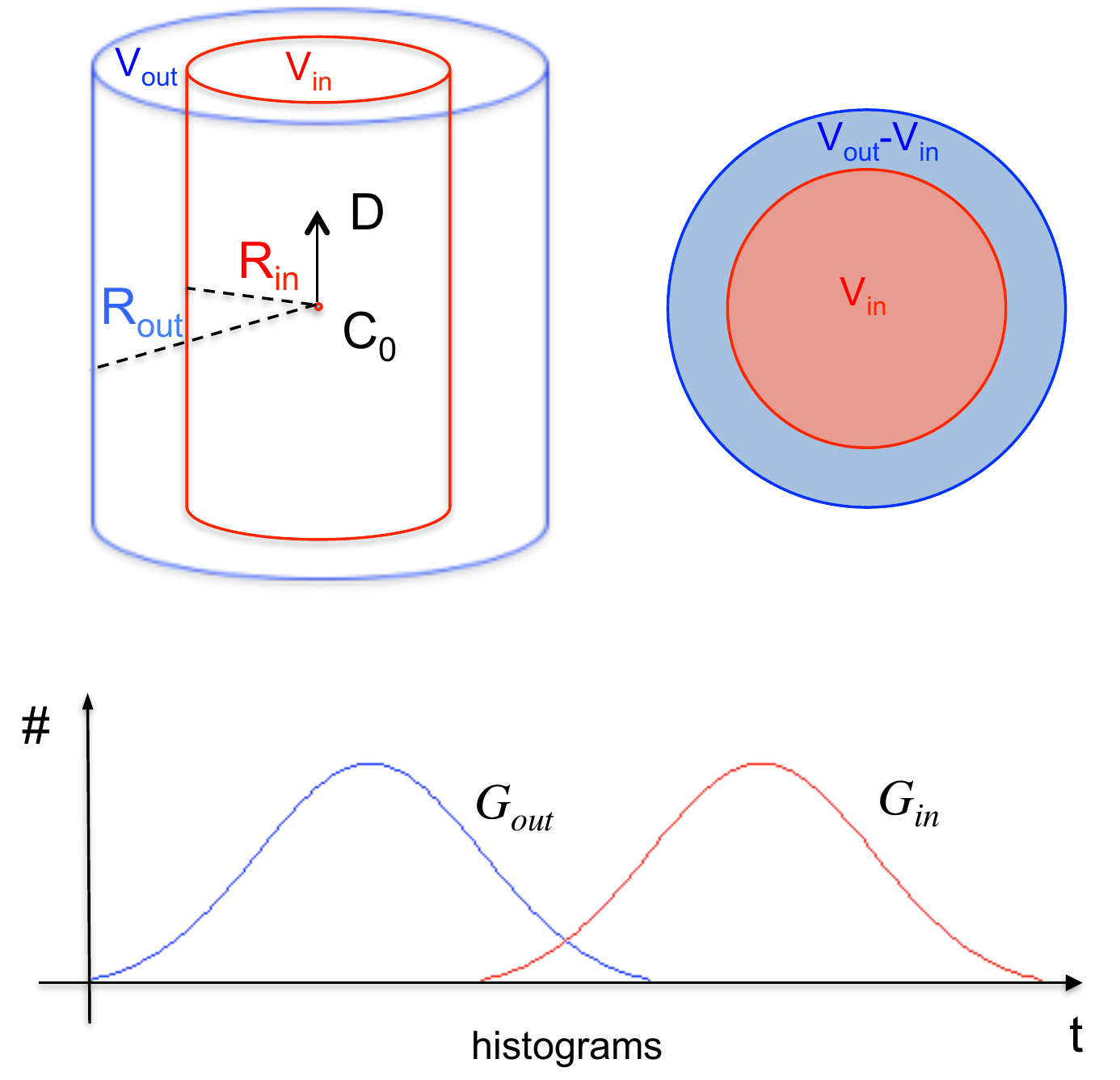}
\caption{Gaussian mixture for cylinder fitting. The red cylinder fits the vessel at a given $\bold{c}$ point, and the blue volume models the neighbouring area.}
\label{fig:gaussian}
\end{figure}

In this work, we propose a vesselness measurement consisting of two concentric cylinders $V_{in}$ and $V_{out}$, comprising respectively a vessel section and its outer neighborhood, as shown in figure \ref{fig:gaussian}. Let $R_{in}$ and $R_{out}$, be the radii of $V_{in}$ and $V_{out}$, respectively, $\bold{c}$ the common central point and $D$ the unitary vector representing the orientation of their common central axis. Let also $R_{out}$ be such that the volume of $V_{in}$ is equal to the volume of $V_{out}$ - $V_{in}$. The cylinder height is defined as two times the inner radius, and any value around this does not impact much the final outcome.

It is further assumed that the voxels’ intensities inside and outside the vessel can be appropriately represented by two distinct Gaussian distributions, denoted respectively $G_{in}(I_p)$ and $G_{out}(I_p)$, where $I_p$ stands for the intensity of the voxel at coordinate $p$. The mean and standard deviation of each Gaussian distribution are computed as the mean value and standard deviation of the voxels inside each corresponding volume $V_{in}$ and $V_{out}$ - $V_{in}$.

The vesselness is then given by:
\small
\begin{align}
\label{eq:vesselness}
W_m =  \min_{R_{in},D,\bold{c}} \frac{\sum_{p \in V_{in}} G_{out} (I_p)  + \sum_{p \in (V_{out}-V_{in})} G_{in} (I_p)}{\|V_{out}\|}
\end{align}
\normalsize


The optimization  implied in the vesselness computation can be performed by  
different stochastic methods. In this work we used Differential Evolution  \cite{diffev} for this purpose.

As byproducts of vesselness computation we obtain the optimum position $\bold{c}$ , the radius $R_{in}$ and the direction $D$ of the hypothesized vessel section. In particular, the optimization procedure starts at a initial location for the vessel point and looks for the most accurate position in its neighborhood. 

\subsubsection{Vessel Point Candidate Selection}
\label{sec:method:detection:selection}

Vessel point candidates are selected from sample points in two steps: 
\begin{enumerate}
	\item Vessel candidates with intensity mean value differing greatly from the mean intensity  value observed at the cone apex point are discarded and not even submitted to the next step, to save processing time.  
	\item Points selected in step 1 are submitted to vesselness computation. The optimal cylinder found by the implied optimization process is validated according to the following rules:
\begin{enumerate}
	\item The mean intensity value of the inner cylinder must be higher than the mean intensity value of the outer volume (see figure \ref{fig:gaussian} for details). This rule models the assumption that vessels appear as roughly bright cylindrical structures in CT images. This rule is applied during the cylinder fit optimization process.
	\item The vesselness of a selected sample point must be higher than a fraction of the vesselness at the seed point. The variable controlling this proportion is user defined and works as a sensibility parameter, which allows for finding weaker vessels (at the cost of adding noise) or just the vascular branches with stronger vesselness information.
	\item Vessel points with too small cylinder radius are discarded, assuming these cases are noise. The threshold for the radius is configurable and set as the image spacing by default (vessels with radius lower than image spacing are hardly detectable).
\end{enumerate}
\end{enumerate}

Having detected the vessel points among sampled points, we structure them as a directed graph as detailed in section \ref{sec:method:sampling} and analyze it for tracking the vascular network, as explained in the next section. 

\subsection{Vascular Network Identification}
\label{sec:method:mapping}

The vascular network tracking for a given collection of vessel point detections is composed of two steps: vascular network tracking and vascular network validation.

\subsubsection{Vascular Network Tracking}
\label{sec:method:mapping:detection}

The idea for detecting vascular networks is to build and analyze local graphs for which the nodes represent vessel detections (as found in Section \ref{sec:method:detection}), using the sample points coordinates and their respective vesselness value. These nodes are connected to neighboring nodes both from adjacent layers (connected by transition edges) and from the same layer (connected by toll edges). Each edge represents the relation between two nodes using a determinate cost, as explained further in this section. 

Thereby, the matching problem can be approached as a minimum-cost network flow problem: finding the optimal set of vessels can be solved by sending flow through the graph so as to minimize its total cost. 

\begin{figure}[t]
\centering
\includegraphics[scale=0.4]{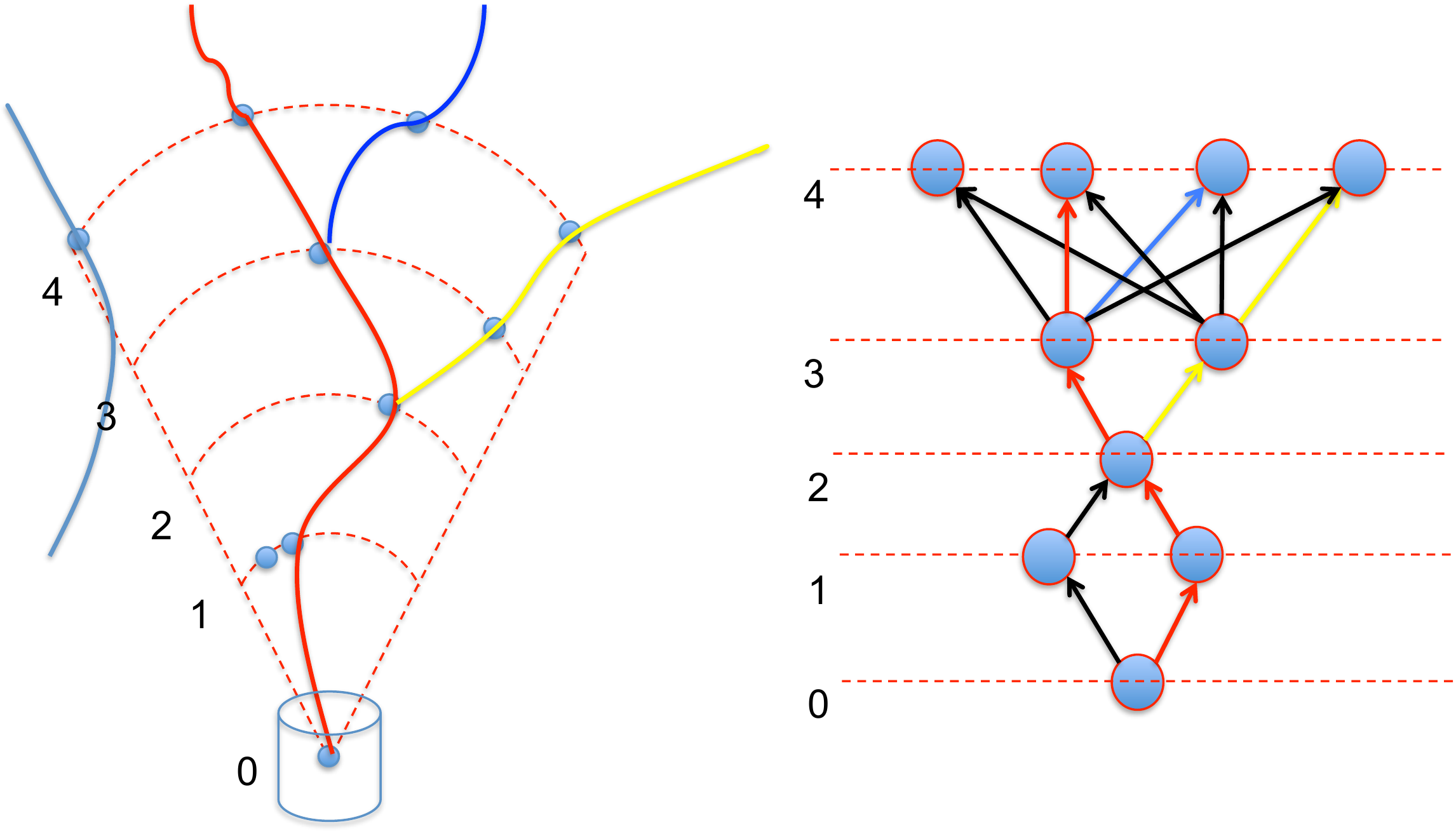}
\caption{Vessel point detections are structured as a directed graph that allows finding vascular network branches, as shown in different colours.}
\label{fig:tracking}
\end{figure}

\normalsize

Let $\mathcal{O}=\{{\bf o}_i\}$ be a set of vessel detections comprehended by the conical sampling volume associated to the current seed (recall Section \ref{sec:method:detection}) whereby ${\bf o}_i=(x_i,y_i,z_i)$   represented by its 3D coordinates. A vessel branch that emanates from the current seed brings about a set of vessel detections  $V_k=\{{{\bf o}_{k_1}},{{\bf o}_{k_2}}, \cdots , {{\bf o}_{k_h}}\}$, laying on a sequence of layers  $k_1, k_2,\cdots k_h$, where $k_i$ is a natural number and $|k_{i+1} - k_i| \leq 1 \quad \textrm{for all} \quad i = 1,...,h-1$, as seen in Figure \ref{fig:conical}. Let $\mathcal{V}=\{V_k\}$ be an arbitrary set of vessel branches. Then, the set of vessel branches $\mathcal{V}*=\{V_k\}$ composing the vascular network can be defined as the set of vessel branches that  best explains the detections in  $\mathcal{O}$. Thus, $\mathcal{V}*$ can be found by maximizing the posterior probability of $\mathcal{V}$ given the set of vessel detections $\mathcal{O}$. 

Assuming that the detections are $i.i.d.$, the objective function is expressed as:

\small
\begin{align}
\mathcal{V}*=\underset{\mathcal{V}}{\operatorname{{\bf argmax}}} \ P(\mathcal{V}|\mathcal{O}) = \underset{\mathcal{V}}{\operatorname{{\bf argmax}}} \ \prod\limits_i P_\textrm{det}({\bf o}_i | \mathcal{V}) P(\mathcal{V}),
\label{eq:mapping}
\end{align}

\normalsize

where $P_\textrm{det}({\bf o}_i | \mathcal{V})$ is the likelihood of $k_1, k_2,\cdots k_h$. 

In order to reduce the space of $\mathcal{V}$, we make the assumption that vessel branches do not overlap, i.e., a detection cannot belong to more than one vessel). Assuming that vessel branches are $i.i.d.$ as well, leads to the decomposition:


\small
\begin{align}
\label{eq:mapfinal}
P(\mathcal{V})&= \prod_{V_k \in \mathcal{V}} P(V_k ) = \prod\limits_{V_k \in \mathcal{V}} P_\textrm{in}({\bf o}_{k_1}) \ldots P_\textrm{t}({\bf o}_{k_i}|{\bf o}_{k_{i-1}})   \ldots P_\textrm{out}({\bf o}_{k_N}) 
\end{align}
\normalsize

for each vessel $\{V_k\}$, represented by an ordered chain. $P_\textrm{in}({\bf o}_{i})$ and $P_\textrm{out}({\bf o}_{i})$ is the probability that a vessel branch starts and ends at detection ${\bf o}_{i}$. $P_\textrm{t}({\bf o}_{i}|{\bf o}_{j})$ is the probability that ${\bf o}_{j}$ is followed by ${\bf o}_{i}$ in the vessel branch. 



\normalsize
To solve the objective function implied in equation \ref{eq:mapping}, we linearize it using the network flow paradigm, where probabilities of connected nodes are represented by means of costs and flows. According to this paradigm, maximizing the probabilities of nodes in a directed network is equivalent to finding the flows that minimize the total flow over the network, given a set of predefined costs. 

\begin{figure}[t]
\centering
\includegraphics[scale=0.4]{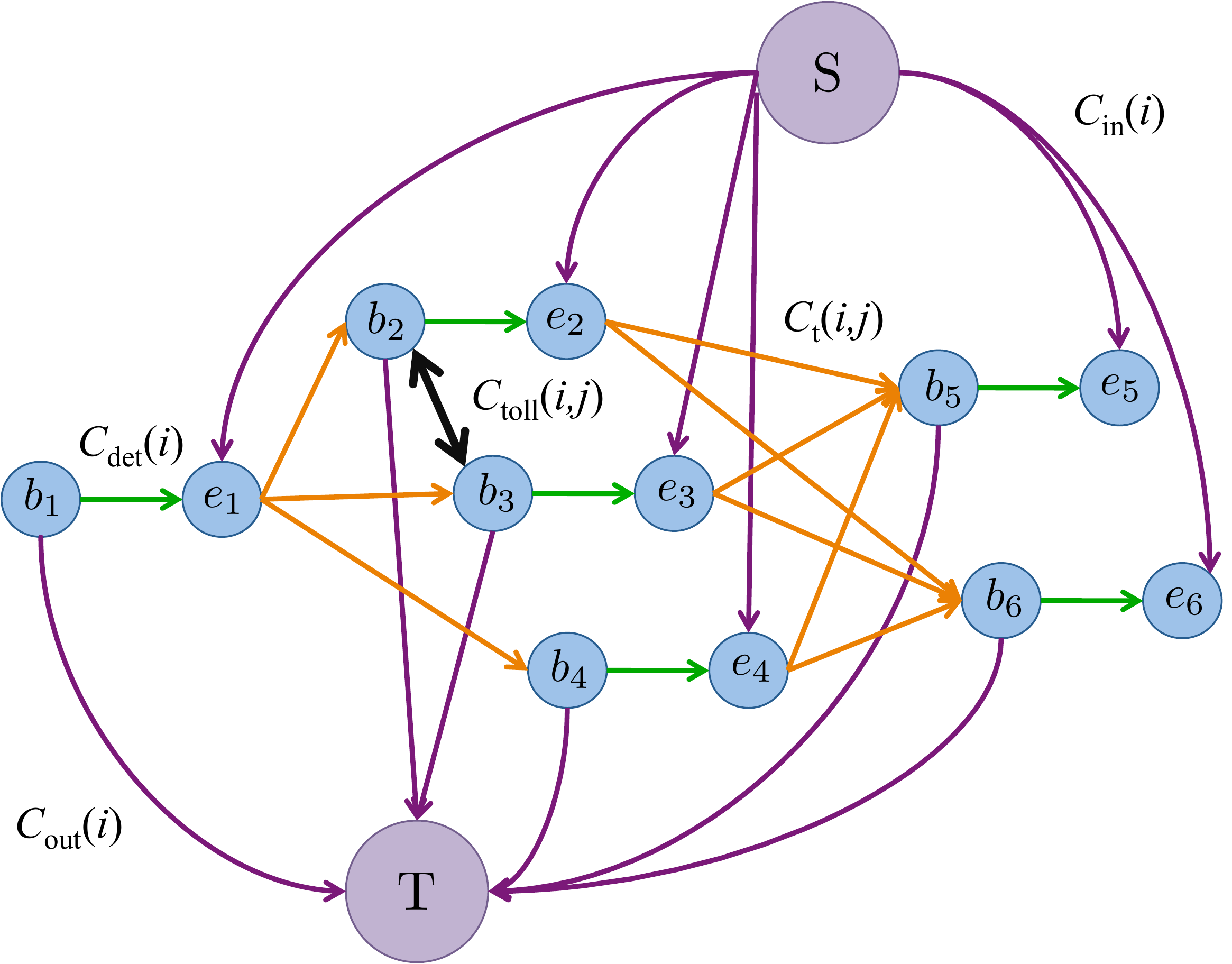}
\caption{Proposed graph structure. First layer of the cone consists of node 1, second layer by nodes 2,3,4 and third layer by nodes 5,6. Each pair of begin/end represents a node referenced by the index associated.}
\label{fig:graph}
\end{figure}

Let $G=(\mathcal{N},\mathcal{E})$ be a directed network with costs associated to every edge $e_i\in \mathcal{E}$. An example of such a network is shown in Fig. \ref{fig:graph}. It contains two special nodes, the source $S$ and the sink $T$. In our proposition, all flow that goes through the graph starts at the $S$ node and ends at the $T$ node, and each flow represents a vessel candidate $V_k$. Each vessel detection ${\bf o}_i$ is represented by two nodes, the beginning node $b_i \in \mathcal{N}$ and the end node $e_i \in \mathcal{N}$ (see Fig. \ref{fig:graph}), and a detection edge connecting them. 

More specifically, each vessel point detection is represented in the graph by a pair of nodes $b_i$ and $e_i$ connected by an edge with cost $C_{det}(i)$ associated to the vesselness measurement for the sampled point. Also, a pair of nodes ($e_i$, $b_i$) representing a vessel point is connected to the vessel points laying in the next layer, say ($e_j$, $b_j$), by so called transition edges with cost $C_t(i,j)$ related to the Euclidean distance between points $i$ and $j$. In this model the optimal path corresponds to the minimum cost, which combines vesselness and the distance between nodes along the path.

Below we detail four of the five different types of edges present in the graphical model. The edges associated to $C_{toll}(i,j)$ are going to be explained later. \\

\normalsize

\noindent{\bf Transition edges.} Transition edges connect end nodes $e_i$ over a given layer to beginning nodes $b_j$ of the next layers (orange edges in Fig. \ref{fig:graph}), with cost $C_{\textrm{t}}(i,j)$ and flow $f_{\textrm{t}}(i,j)=1$, if ${\bf o}_i$ and ${\bf o}_j$ belong to $V_k$, and $0$ otherwise.
The cost associated to transition edges relates to the distance of adjacent vessel detections. Since nearby points are more likely to belong to the same vessel, we define the costs to be a decreasing function of the distance between neighboring vessel detections, assuming a distance $D_{\textrm{max}}$:


\small
\begin{align}
C_\textrm{t}(i,j)&= -\log \left( \frac{D_{\textrm{max}}- \|({\bf o}_j - {\bf o}_i)\|}{D_{\textrm{max}}} \right) 
\label{linkedge}
\end{align}
\normalsize

\noindent{\bf Detection edges.} These edges (plotted in green in Fig. \ref{fig:graph}) connect the beginning node $b_i$ and end node $e_i$, with flow $f_{\textrm{det}}(i)=1$ if ${\bf o}_i$ belongs to $V_k$, and $0$ otherwise. 
It is worth noting that, if all edge costs are positive, the solution of the minimum-cost problem is the trivial null flow. The trick in our model is to represent each vessel detection ${\bf o}_i $ with two nodes and a detection edge in between with negative cost $C_{\textrm{det}}(i)=\log \left(1-P_\textrm{det}({\bf o}_i) \right)$.
The higher the likelihood of a detection $P_{\textrm{det}}({\bf o}_i)$ the more negative the cost of the detection edge. Hence true detections are likely to be in the path of the flow so as to minimize the total cost.\\


\noindent{\bf Entrance and exit edges.} Entry edges (purple in Fig. \ref{fig:graph}) connect the source $S$ to all end nodes $e_i$, with cost $C_{\textrm{in}}(i)=1$ and flow $f_{\textrm{in}}(i)$. Similarly, exit edges connect the start node $b_i$ with sink $T$, with cost $C_{\textrm{out}}(i)=1+C_{\textrm{det}}(i)$, to ensure that a trajectory ends at a detection with high probability, and flow $f_{\textrm{out}}(i)$. The flows are 1 if the trajectory $V_k$ starts/ends at ${\bf o}_i$. It is interesting to notice that these constants were arbitrarily set, since the flow optimization would find the same paths independently on their specific values.\\





Now equation \ref{eq:mapping} can be rewritten in terms of costs and flows, as usual in the network flow paradigm. First, we apply negative log-likelihoods to the equation \ref{eq:mapping} to derive equation \ref{eq:LP1}:

\small
\begin{align}
\label{eq:LP1}
\mathcal{V}*&=\underset{\mathcal{V}}{\operatorname{{\bf argmin}}} [-\log P(\mathcal{V})   + \sum_{i} -\log P_\textrm{det}({\bf o}_i | \mathcal{V})]  
\end{align}
\normalsize

then, denoting each of the probability terms in equations \ref{eq:LP1} and \ref{eq:mapfinal} as a product of costs and flows, we derive equation \ref{eq:LP}, which computes the set of flows $\mathcal{F}*$ that minimizes the global network flow. Since maximizing the probabilities in the graph is equivalent to finding the nodes that minimize the total flow over the network, we find $\mathcal{V}*$ indirectly: the set of $V_k$ vessels of the optimal vascular network $\mathcal{V}*$ corresponds to the set of paths of minimum flow in $\mathcal{F}*$. All the related symbols are intrinsically related to the graph shown in Figure \ref{fig:graph}. 


\small
\begin{align}
\label{eq:LP}
\mathcal{F}*&=\underset{f_{\textrm{in}},f_{\textrm{t}},f_{\textrm{out}},f_{\textrm{det}}}{\operatorname{{\bf argmin}}}  [\sum_{i} C_{\textrm{in}}(i)f_{\textrm{in}}(i)  + (\sum_{i,j} C_{\textrm{t}}(i,j)  f_{\textrm{t}}(i,j)) + C_{\textrm{out}}(i)f_{\textrm{out}}(i) + C_{\textrm{det}}(i)f_{\textrm{det}}(i)]  
\end{align}
\normalsize

subject to the following constraints: 

\begin{itemize}
\renewcommand{\labelitemi}{$\bullet$}
\item{Edge capacities: we assume that each detection belongs either to one vessel or to none, i.e., flows can assume 0 or 1 values, or $0 \leq f(i)\leq 1$ in its linearly relaxed form.}\\
\item{Flow conservation at the nodes: the entering flow of a node equals its exiting flow.}
\small
\begin{align}
 \label{eq:flow}
\textstyle
 f_{\textrm{in}}(i) + f_{\textrm{det}}(i) =  \sum_{j} f_{\textrm{t}}(i,j) \qquad \quad \sum_{j} f_{\textrm{t}}(j,i) = f_{\textrm{out}}(i)+ f_{\textrm{det}}(i) 
\end{align}
\normalsize
\end{itemize}

By now we have fully defined a linear program for finding  $\mathcal{F}*$ and consequently $\mathcal{V}*$. This linear problem can be solved using many of the Linear Programming solvers proposed thus far by the optimization community, such as Simplex or k-shortest paths \cite{LP}. However, equation \ref{eq:LP} is very computationally intensive. 
We propose a fast iterative procedure to compute  $\mathcal{V}*$, which finds the global solution for each  vessel sequentially and also identifies the network topology.



The first vessel $V_1$ is found by solving Eq. \ref{eq:LP}, allowing entering flow only at the seed point (the cone apex) and setting the maximum flow going out of node $T$ to be 1. 
A further condition is imposed in order to avoid multiple paths representing a single vessel. Specifically, the vessel must be distant enough from $V_1$, in order to be included in  $\mathcal{V}*$. 
To represent such condition in the graph structure a new type of edge is proposed, which connects vessel detections. Edges of this type are associated to a penalty cost, hereby called ``toll''. They are represented by the thick black edges in Fig. \ref{fig:graph}. This "toll" cost is defined as:

\small
\begin{align}
\label{eq:toll}
C_\textrm{toll}(i,j)=K_\textrm{toll} \cdot \exp \left( \frac{-\|{(\bf o}_i - {\bf o}_j)\|}{D_\textrm{radius}} \right) 
\end{align}
\normalsize
where $K_\textrm{toll}$ stands for a weight, and ${D_\textrm{radius}}$ stands for a maximum penalty distance in millimeters.

For all vessel detections ${\bf o}_j$ which are at a distance ${D_\textrm{radius}}$ or less to any vessel detection found thus far in vessel branch $V_1$, we compute a corresponding $C_\textrm{toll}$, which will be included to Eq. \ref{eq:LP} as a positive penalty. We privilege this way the detection of branches other than $V_1$, since too similar paths will be  penalized through the summation of toll costs. We found out empirically that W $K_\textrm{toll}=5$ is a proper choice.

\begin{figure*}[t]
\centering
\subfigure[]{
\includegraphics[width=0.2\linewidth]{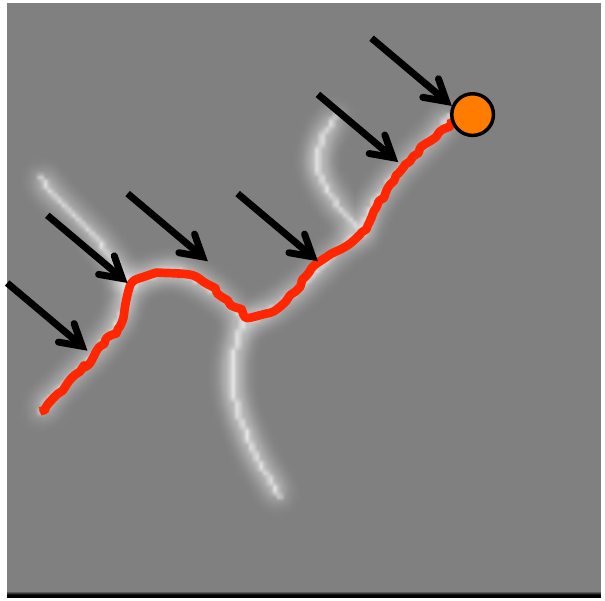} 
\label{prob1}
}
\subfigure[]{
\includegraphics[width= 0.2\linewidth]{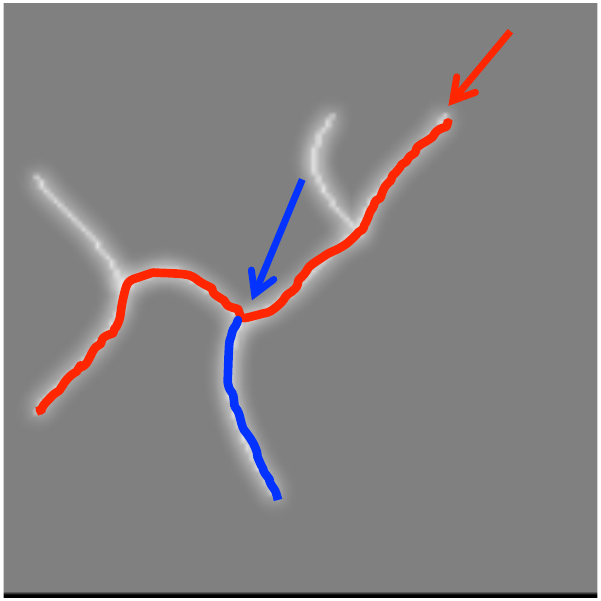} 
\label{prob2}
}
\subfigure[]{
\includegraphics[width= 0.2\linewidth]{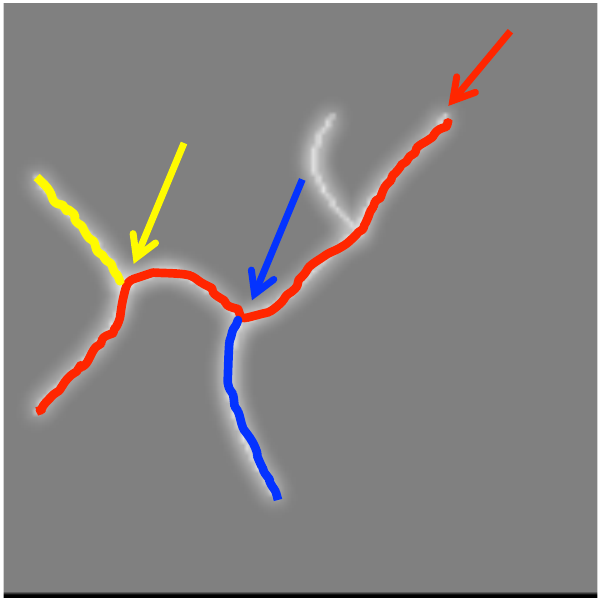} 
\label{prob3}
}
\subfigure[]{
\includegraphics[width= 0.2\linewidth]{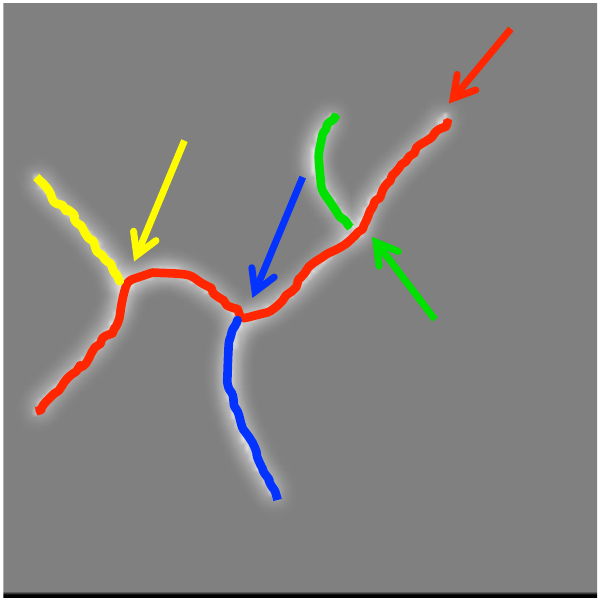} 
\label{prob4}
}
\subfigure[]{
\includegraphics[width= 0.23\linewidth]{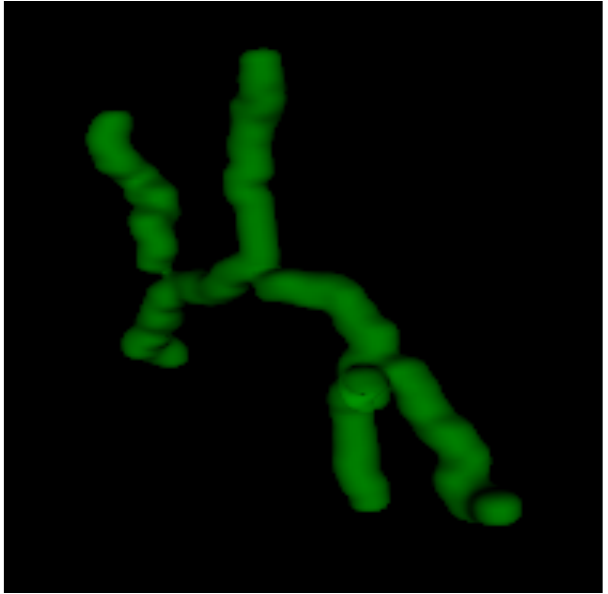} 
\label{prob5}
}
\vspace{-0.3cm}
\caption{Proposed optimization method example of synthetic images. (a) Initial point marked in orange. First path found is the one with minimum cost (red). In the next step, sources will be added along the path (black arrows). (b,c,d) Paths found iteratively, sources of each path marked by colored arrows. (e) 3D view of the vascular network.}
\label{fig:syntree}
\vspace{-0.2cm}
\end{figure*}

For detecting bifurcations and their corresponding branches a simple solution is adopted. We update the flow conditions, allowing any point of the detected $V_k$ 
to be the starting point of a new vessel branch, as shown in Fig. \ref{fig:syntree}. Formally this is done by allowing $f_{in}$ to range between {0,1} for such vessel points, and by setting it to zero for all the other vessel detections. This solution handles bifurcations, and new branches are found iteratively. $f_{in}$ and toll costs are updated accordingly, until the cost of finding a new vessel branch becomes positive, i.e., until the positive costs outweigh the negative costs in the minimum flow corresponding to the new path. 

\subsubsection{Vascular Network Validation}
\label{sec:method:post}

\begin{figure*}[t]
\centering
\subfigure[]{
\includegraphics[scale=0.5]{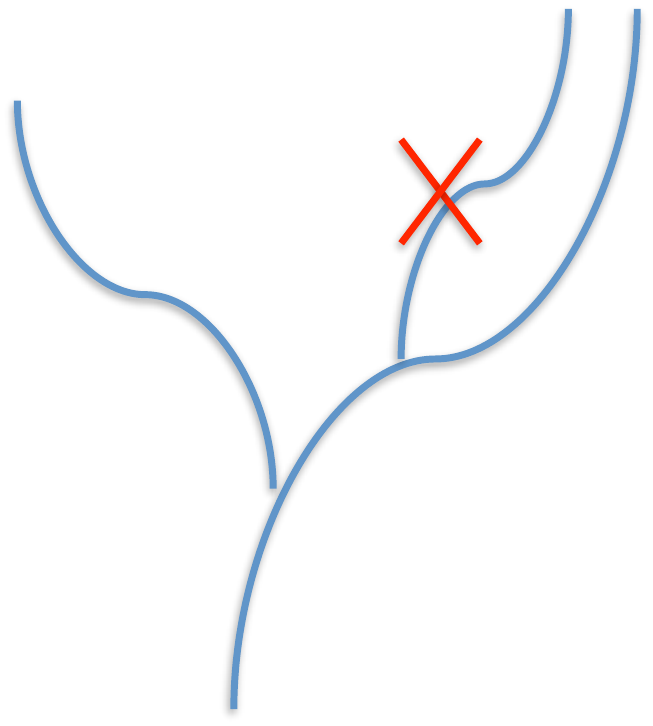} 
\label{post_a}
}
\subfigure[]{
\includegraphics[scale=0.5]{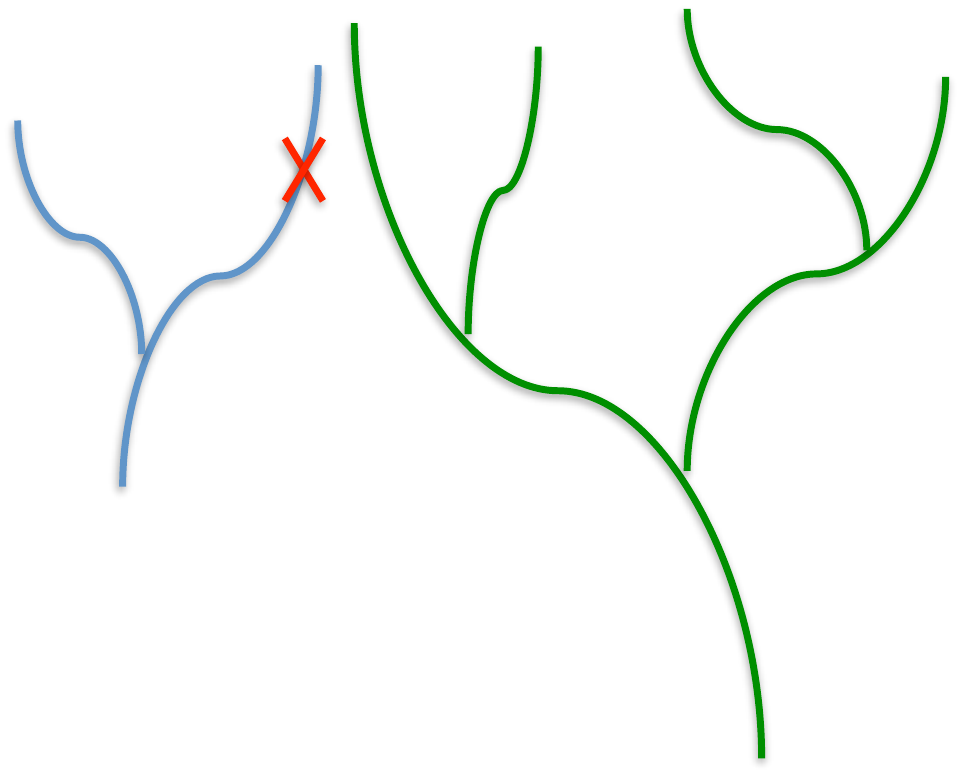} 
\label{post_b}
}
\caption{Post processing impose some anatomical constraints for a vascular network. The red 'X' shows the branches which would be eliminated following each rule (a) and (b) described in section \ref{sec:method:post}.}
\label{fig:post}
\end{figure*}

A final vascular network validation procedure is performed to ensure that the final outcome  (figure \ref{fig:post})  conforms to vascular anatomy, specifically:

\begin{enumerate}
	\item Vessel branches cannot be too close to each other along their length, in which case, the  detected branches are very likely to represent the same vessel. This condition is implemented by imposing a minimum distance between  branches found at each iteration. If this value is lower than a threshold, one of them is discarded.
	\item In order to avoid loops, vessel branches are not permitted to reconnect to the so far segmented vascular network. This is implemented by computing the minimum distance between the branches found and the already segmented network. 
If this value is lower than a threshold, the branch forming a loop is discarded.
\end{enumerate}

It is important to mention that those threshold values are relative to the estimated radius. In this way, the absolute threshold values change dynamically during the segmentation process, so as to avoid misconnections, while still keeping the ability to find small vessels.

\subsection{Next seeds definition}
\label{sec:method:next}

\begin{figure}[t]
\centering
\includegraphics[scale=0.5]{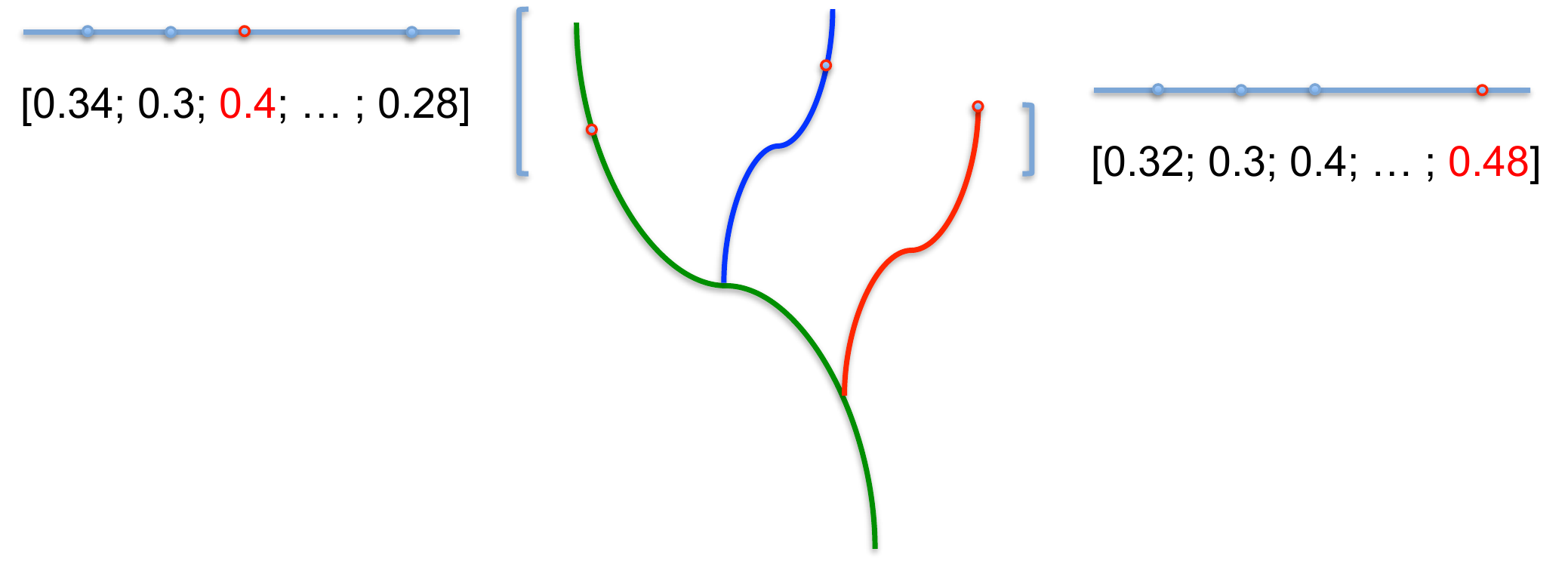}
\caption{Definition of the seeds for the next iteration.}
\label{fig:seed}
\end{figure}

The final step consists in defining new seeds from the detected branches to be used in the iterative method. A simple procedure is defined to find the best seed of each branch: we seek for the vessel point detection with the best vesselness value in the ending part (30\% last points) of each branch found, and define it as a new seed, as depicted in figure \ref{fig:seed}.

Each new seed derives a new vascular network tracking process at its location, until no more seeds are found. An overview of the proposed method is shown in Algorithm \ref{alg1}.

\begin{algorithm}[t] 
\small           
\caption{Iterative vessel network tracking}          
\label{alg1}                           
\begin{algorithmic}                    
\WHILE{$\| \mathcal{S} \| >0$}
\vspace{0.13cm}
\STATE 1. Get a seed from $\mathcal{S}$ and find the vessel direction
\vspace{0.13cm}
\STATE 2. Compute a sampling cloud of points as shown in section \ref{sec:method:sampling}
\vspace{0.13cm}
\STATE 3. Build the graph from the sample points
\vspace{0.13cm}
\STATE 4. Compute vesselness measurements at sample points given Eq. \ref{eq:vesselness}
\vspace{0.13cm}

\WHILE{$ C(V_n)  <0$}
\vspace{0.13cm}
\STATE 5. Find vessel $V_n$ with minimum cost
\vspace{0.13cm}
\STATE 6. Compute the toll charges (Eq. \ref{eq:toll}) and new flow conditions.
\vspace{0.13cm}
\ENDWHILE

\vspace{0.13cm}
\STATE  6. Define new seeds $\mathcal{S}$
\vspace{0.13cm}
\ENDWHILE
\end{algorithmic}
\normalsize
\end{algorithm}

\afterpage{\clearpage}

\section{Results and Discussion}
\label{sec:resultado:resultados}

\subsection{Results}

Experiments have been conducted to evaluate the proposed method. The scarcity of public data sets with reference information has prevented a thorough and objective experimental comparison among (semi) automatic methods for vascular segmentation proposed thus far. In fact, to our best knowledge, for this reason none of these works managed to perform an objective performance comparison with alternative approaches for vessel network segmentation. 

Some authors tried to circumvent this difficulty either by relying on synthetic data (e.g. \cite{frimanmia2010,worz2007,WongC07,schap07}) or by limiting themselves to visual evaluations (e.g. \cite{worz2007,WongC07,Manniesing07,schap07}). The analysis reported hereafter follows one or the other strategy depending on the characteristics of the test data. 

In some experiments we used a data set created for conference challenges, which contain references. In those cases the following accuracy metrics were used (detailed in \cite{cls2009}):
\begin{itemize}
\item The overlap Dice similarity index for 3D volumes.
\item Root mean squared (RMS) distance between reference and segmented 3D surfaces.
\item Hausdorff distance between reference and segmented 3D surfaces.
\end{itemize}

Concerning the processing time, it is important to mention that the iterative nature of the method can derive very variable amount of time to run depending on the complexity of the segmented network. Just to give an idea of processing time, an iteration with 5000 points can take between 5 and 10 minutes approximately. Depending on the complexity of a network, the number of cones needed in the iterative process to segment the whole network vary a lot. 

It is worth mentioning that the optimization procedure implied in the vesselness computation was done using the differential evolution \cite{diffev} method.

In the following we report the experiments conducted to evaluate the proposed method. 

\subsubsection{Synthetic data 1}

\begin{figure}[t]
\centering
\includegraphics[scale=0.6]{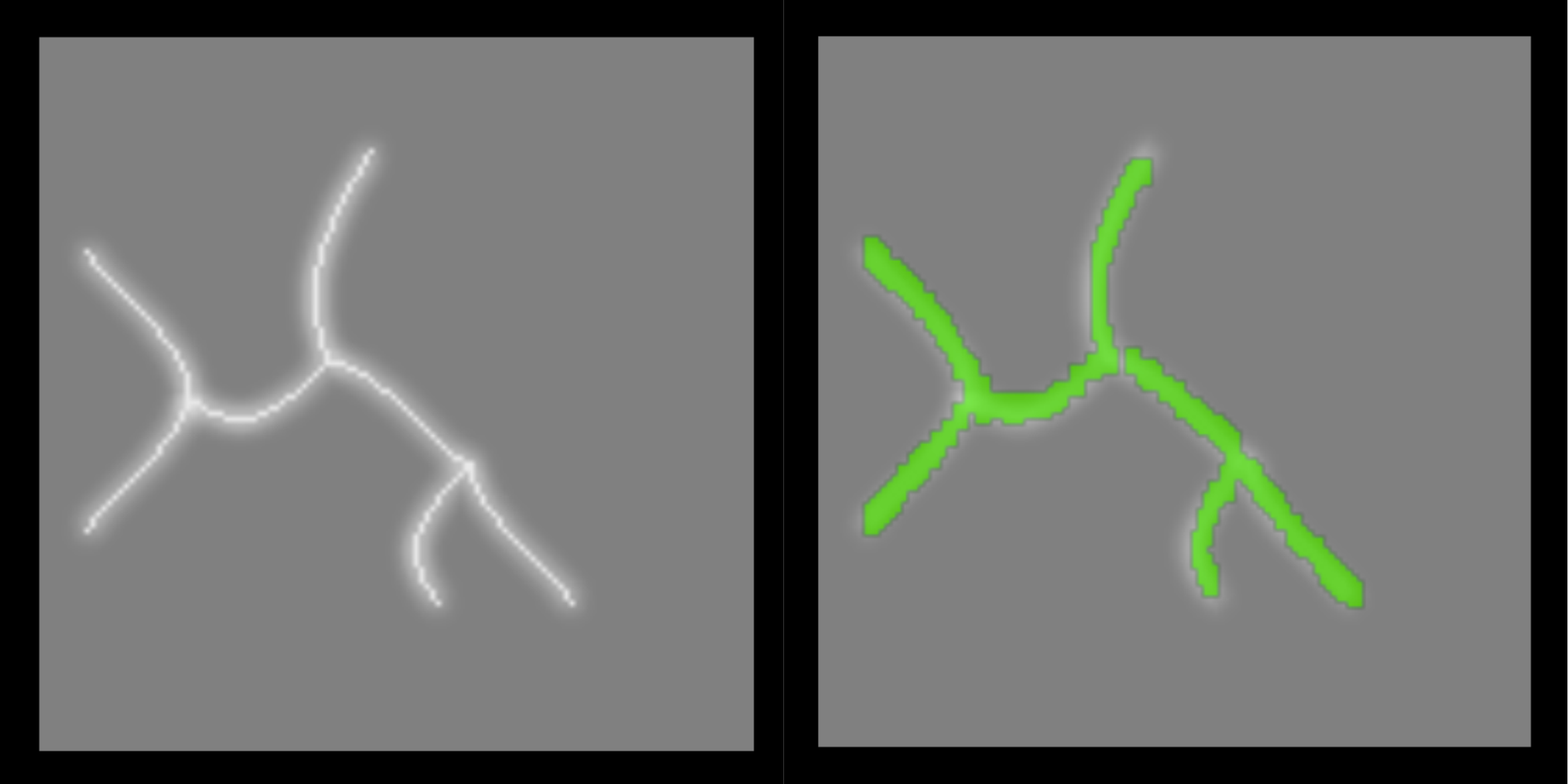}
\caption{Synthetic data segmentation using sinusoidal shaped vascular-like structures. On the left side the input data; on the right side the segmented vascular network in green.}
\label{fig:result1}
\end{figure}

The first experiment was based on a data set available in \cite{Macedo10}. It consists of  a simple planar vascular like structure modeled as sinusoidal shapes with bifurcations.  The data set is affected by Gaussian noise artificially added to the raw data.The result obtained with this data-set is shown in figure \ref{fig:result1}. Clearly, the vascular network was fully segmented. Even though no reference is available, a visual inspection  indicates that the method was successful in this case.

\subsubsection{Synthetic data 2}

In these experiments a more complex synthetic data-set provided by \cite{MiguelGalarreta2012mastery, Miguel2013Vessels} has been used. The data-set of a network of three-dimensional synthetic blood vessels generated by so called stochastic Lindenmayer systems
(L-systems) relying on grammars that represent blood vessel architectures so as to produce  nearly realistic vascular networks. 
Nine different sequences were used in our experiments, each of them with different morphological characteristics. Our method achieved maximum accuracies ranging from 98\% to 100\% according to the overlap Dice similarity index described in \cite{cls2009}.

\begin{figure*}[t]
\centering
\subfigure[]{
\includegraphics[width=0.45\linewidth]{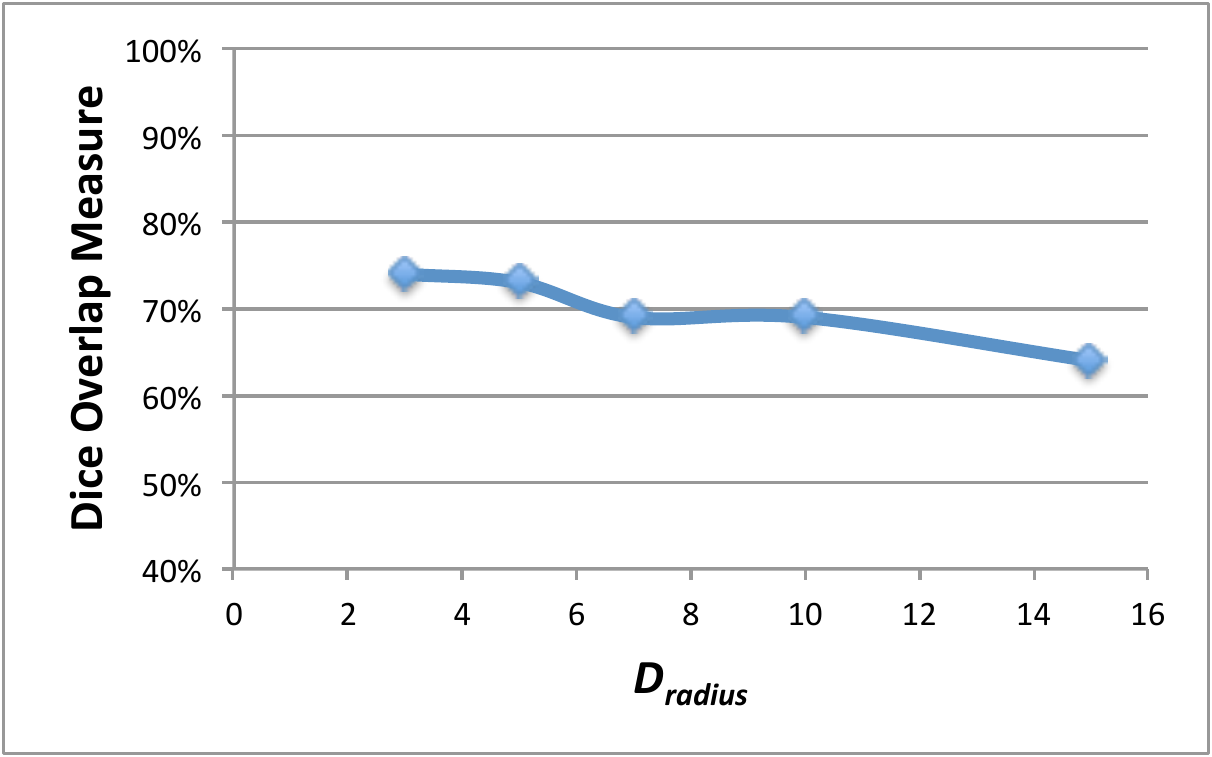} 
\label{param1}
\hspace{-0.15cm}
}\
\subfigure[]{
\includegraphics[width= 0.45\linewidth]{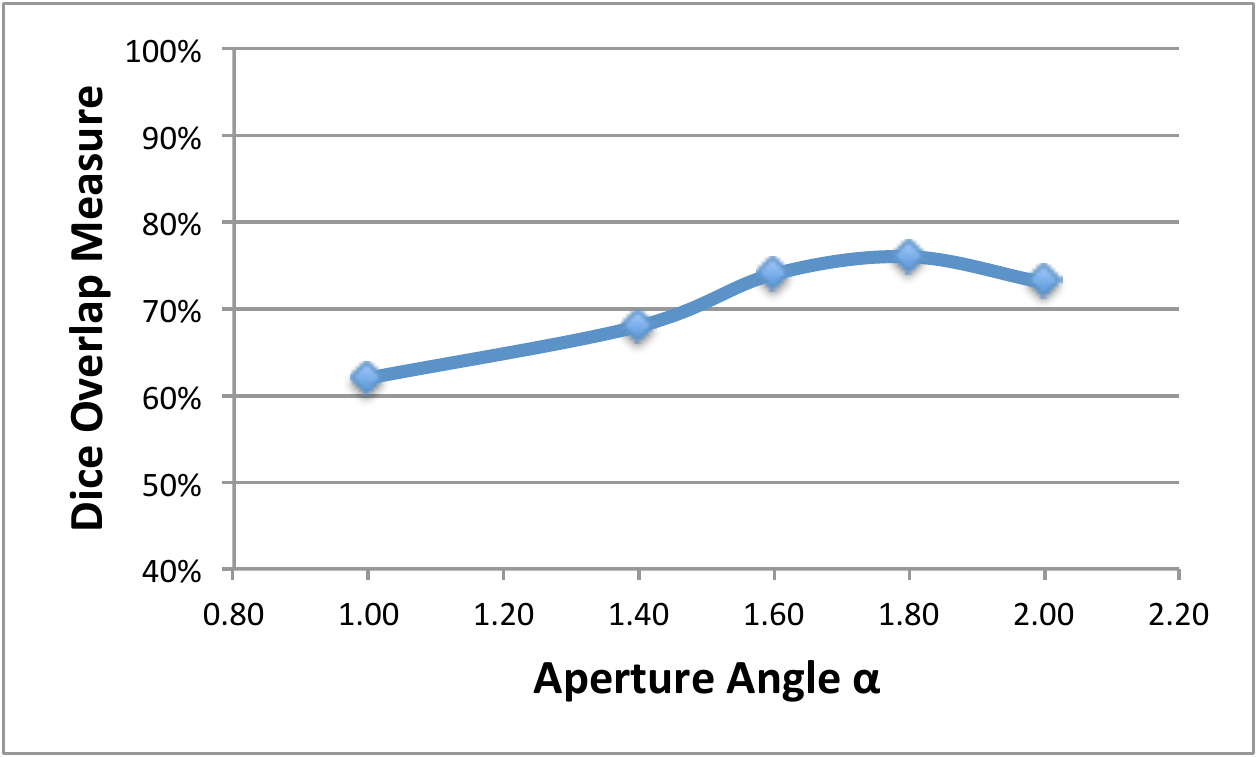}  
\label{param2}
\vspace{0.15cm}
}
\\
\subfigure[]{
\includegraphics[width= 0.45\linewidth]{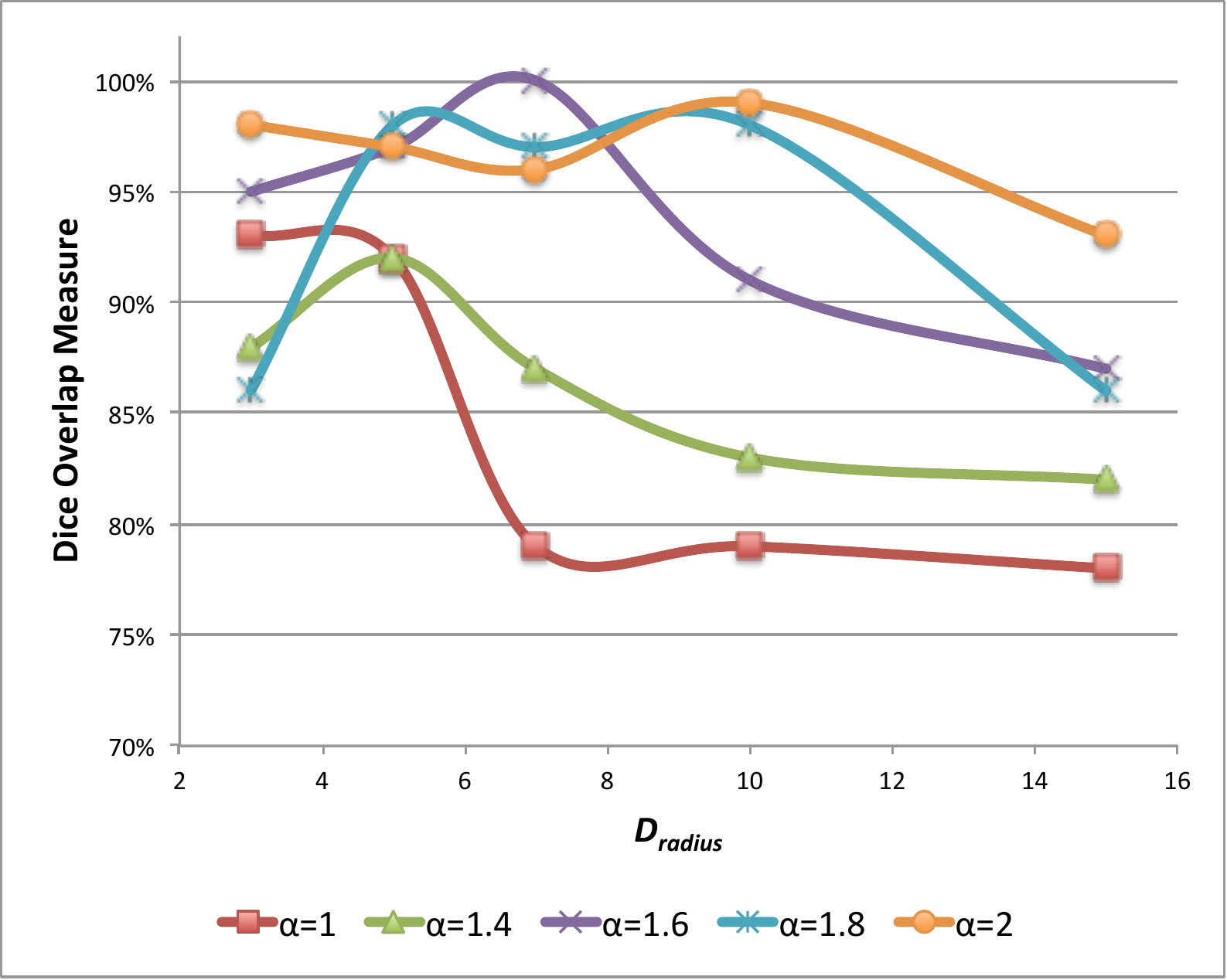} 
\label{param3}
\hspace{-0.15cm}
}\
\subfigure[]{
\includegraphics[width= 0.45\linewidth]{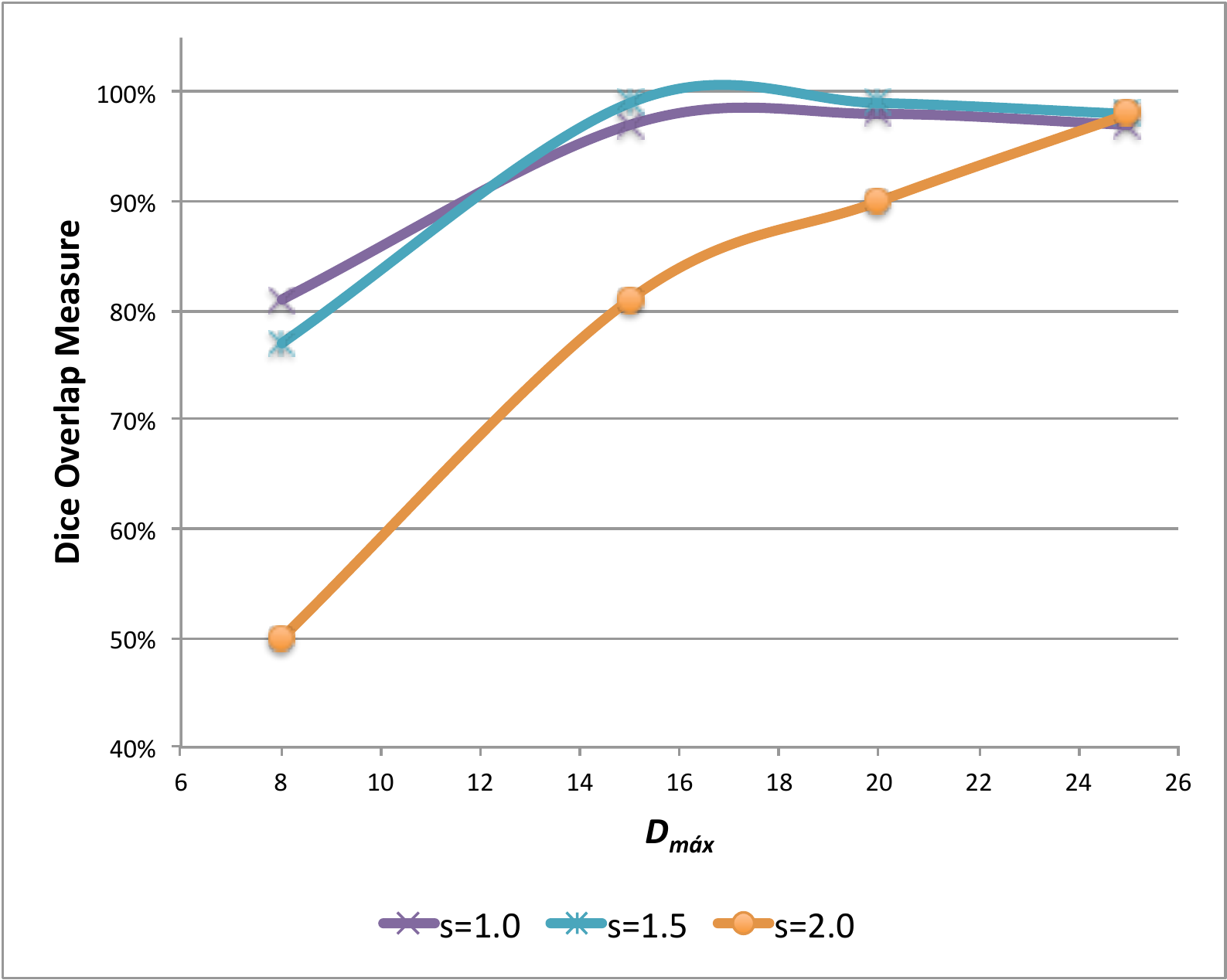} 
\label{param4}
}
\vspace{-0.3cm}
\caption{Parameter study of $D_\textrm{radius}$ in mm, $D_\textrm{max}$ in mm, aperture angle $\alpha$ in radians and sampling distance $s$ in mm (see sections \ref{sec:method:sampling} and \ref{sec:method:mapping}). In each graph we see the performance achieved using different pairs of parameters values.}
\label{fig:params}
\vspace{-0.4cm}
\end{figure*}

Figure \ref{fig:params} shows the results of  experiments to assess the sensibility of our method to its parameters.Two of them were the sampling distance ($s$) and aperture angle ($\alpha$) of the conical
sampling cloud described in section \ref{sec:method:sampling}. Two other parameters related to the Linear Programming
optimization described in section \ref{sec:method:mapping} have also been tested, corresponding to $D_\textrm{radius}$ from equation \ref{eq:toll} and $D_\textrm{max}$ from equation \ref{linkedge}.
We used the overlap Dice similarity index to compare the volume found during the segmentation process and the one of the original image. 

Figure \ref{param1} shows the average overlap measure computed over all nine sequences for varying values of $D_\textrm{radius}$. It is possible to notice that on average $D_\textrm{radius}$ must be kept small (from 3 to 5), otherwise bifurcations are not detected because toll costs are too high, causing the drop in accuracy. In figure \ref{param2} we make a similar study for the aperture angle $\alpha$ parameter. Interestingly, this parameters is related to anatomical characteristics of the vascular network (or part of it) to be segmented, which might spread more or less widely. This fine tuning is necessary because if the angle is too small bifurcations cannot be properly followed because they do not fall inside the cone; on the other hand, if the angle is too large, surrounding structures are included in the conical volume and spurious paths can be found. 

Figure \ref{param3} presents the relationship between $D_\textrm{radius}$, the aperture angle $\alpha$ and the resulting average overlap measure. As the $\alpha$ value used gets higher, the optimum $D_\textrm{radius}$ also increases, so as to avoid that too many false bifurcations are found. Nonetheless, the results are very stable with a wide range of parameters configuration. Finally, the plot of figure \ref{param4} shows the relation between $D_\textrm{max}$ and the sampling distance $s$. Clearly, the optimum value of $D_\textrm{max}$ increases with the sampling distance $s$. This is not unexpected, since $C_t$ is related to the distance of the corresponding vessel points, which are clearly affected by the sampling distance. A sampling distance of 1-1.5 mm gave in our experiments good and stable results. 

\begin{figure}[t]
\centering
\includegraphics[scale=0.6]{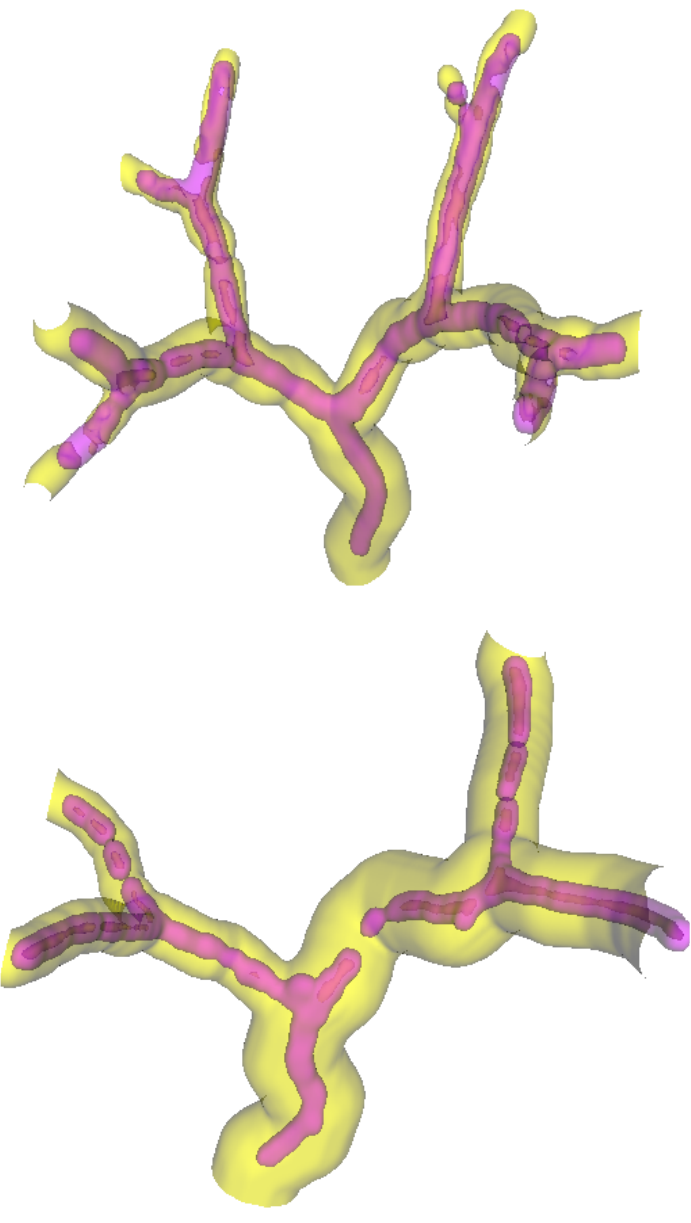}
\caption{Synthetic data segmentation using L-Systems. It is possible to see the reference in yellow and the results achieved in pink. }
\label{fig:estenose}
\end{figure}

Figure \ref{fig:estenose} shows the outcome and provides a visual evaluation of its accuracy. Since the paths are composed by sampling points, the pink part was artificially filled, and this is the reason of the split observed in the bottom model.

\subsubsection{Pulmonary Data}

This dataset was provided by the Extraction of Airways from CT 2009 (EXACT09) challenge \cite{pulmonary} and contains pulmonary vessels inside the lungs.

The goal of this challenge was to compare the results of various algorithms to extract the airway tree from CT scans using a common dataset and performance evaluation procedure. The challenge provides training and testing datasets, and there is also a reference data available for the pulmonary airways but not for the pulmonary vascular network, so, our evaluation in this case was qualitative. 

\begin{figure}[t]
\centering
\includegraphics[scale=0.5]{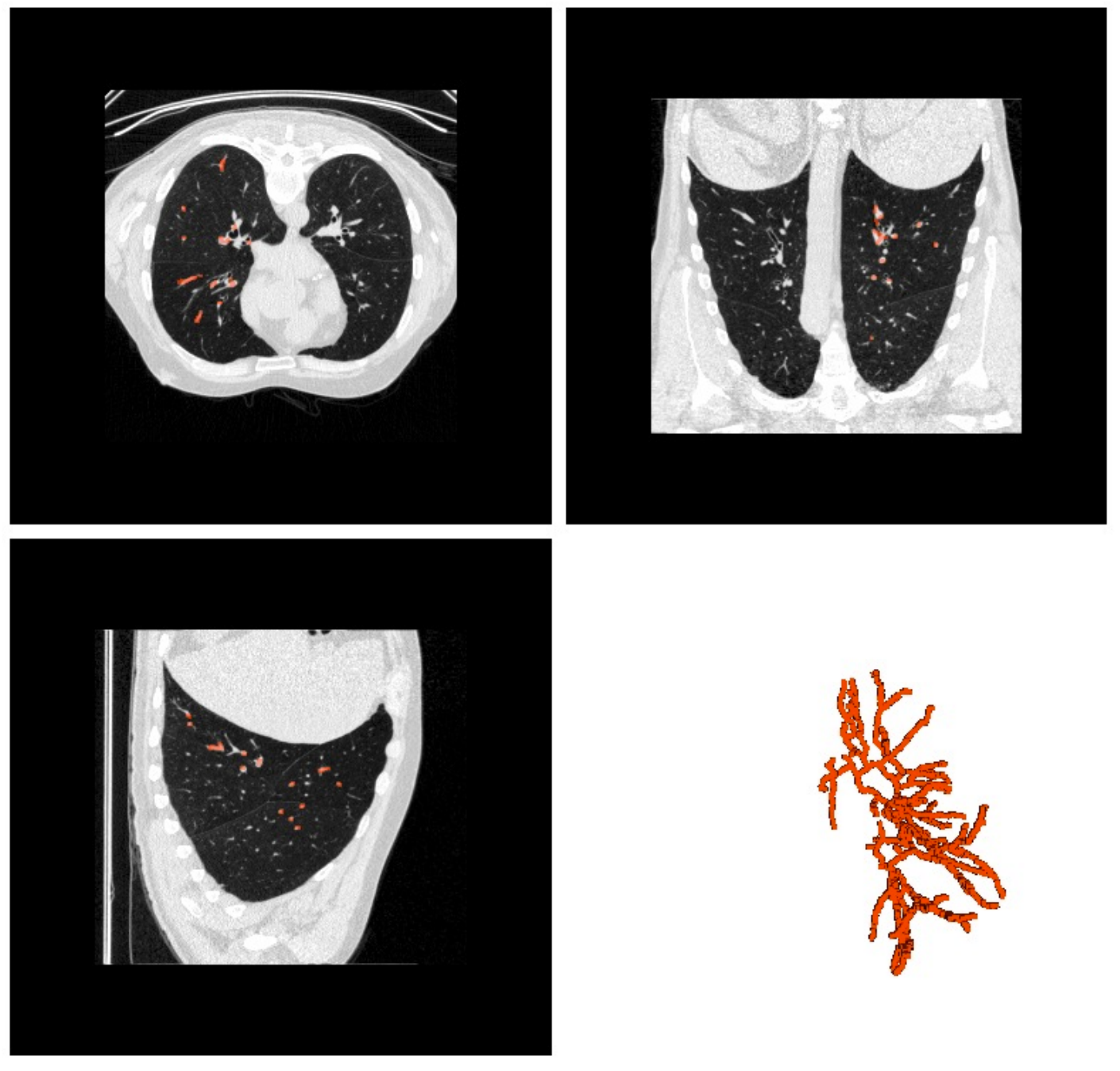}
\caption{Vascular segmentation for pulmonary real dataset. We see the results in the axial, coronal and sagittal views and the 3D model generated.}
\label{fig:pulmonary}
\end{figure}

Figure \ref{fig:pulmonary} shows the results produced by our method on this data set. It can be seen that most of the vascular network, including bifurcations, were found. Even though this kind of evaluation is not numerical, the results show the potential of the method, specially if one takes into consideration that a single seed point was used.  

\subsubsection{Coronary Data}

In this experiment we tested our method for cardiac vessels segmentation. To evaluate our algorithm for such application, we used the coronary dataset provided for  the MICCAI "3D Segmentation in the Clinic: A Grand Challenge II” \cite{coronary}. 
This database contains thirty-two cardiac CT datasets with reference data available for the four main coronary vessels. It is important to notice that the reference is composed by four single vessel segmentation, and therefore does not test the full potential of our method since they do not form a full vascular network. Nonetheless, it is a very interesting dataset and the available references can be used as guidance for visual assessment.  

\begin{figure}[t]
\centering
\includegraphics[scale=0.5]{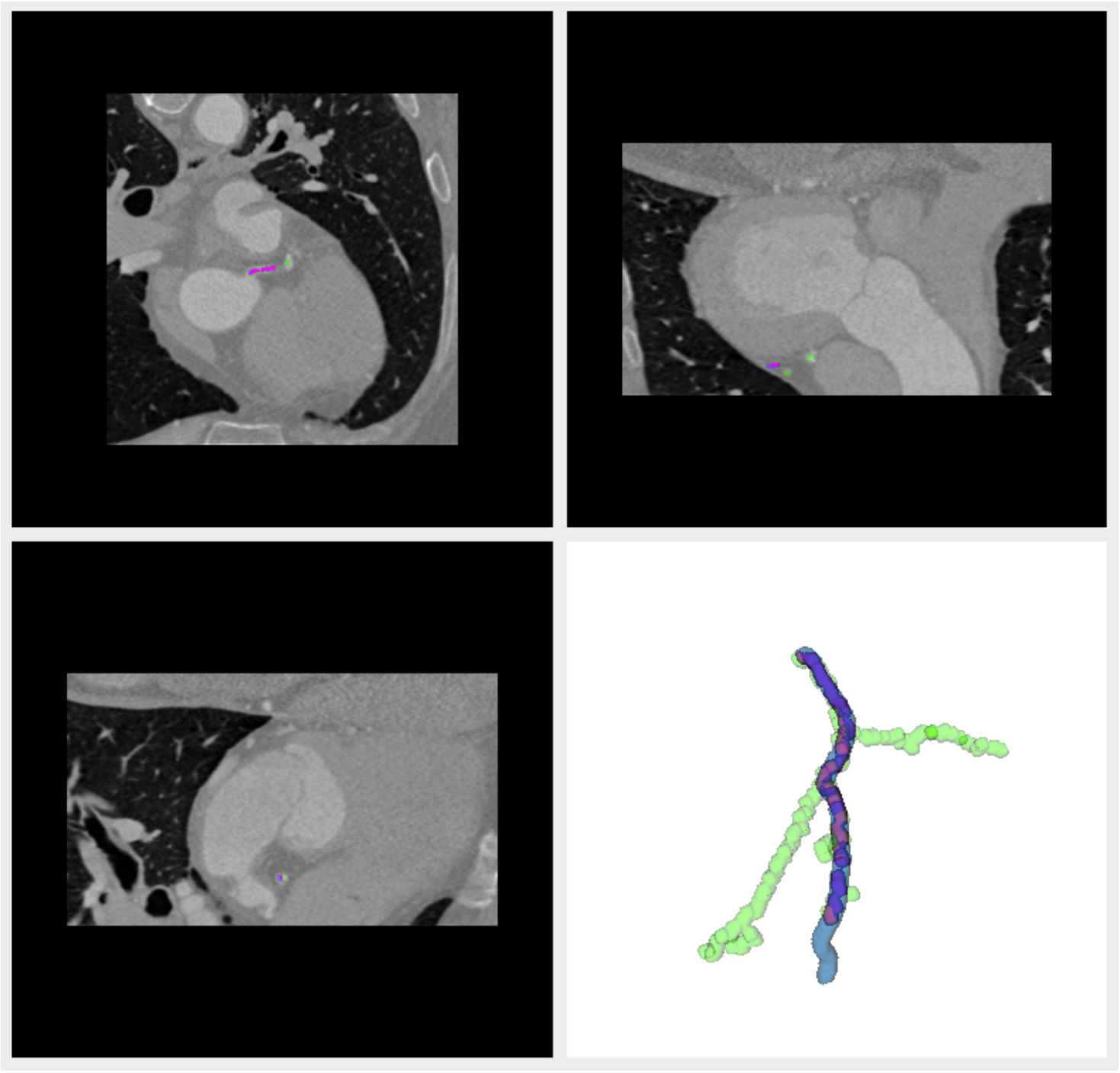}
\caption{Vascular segmentation for coronary real dataset. We see the results in the axial, coronal and sagittal views and the 3D model generated.}
\label{fig:coronary}
\end{figure}

Figure \ref{fig:coronary} shows that the reference single vessel is among the branches segmented (in green) using our method, which segmented other two extra branches as well.  

\begin{figure}[t]
\centering
\includegraphics[scale=0.5]{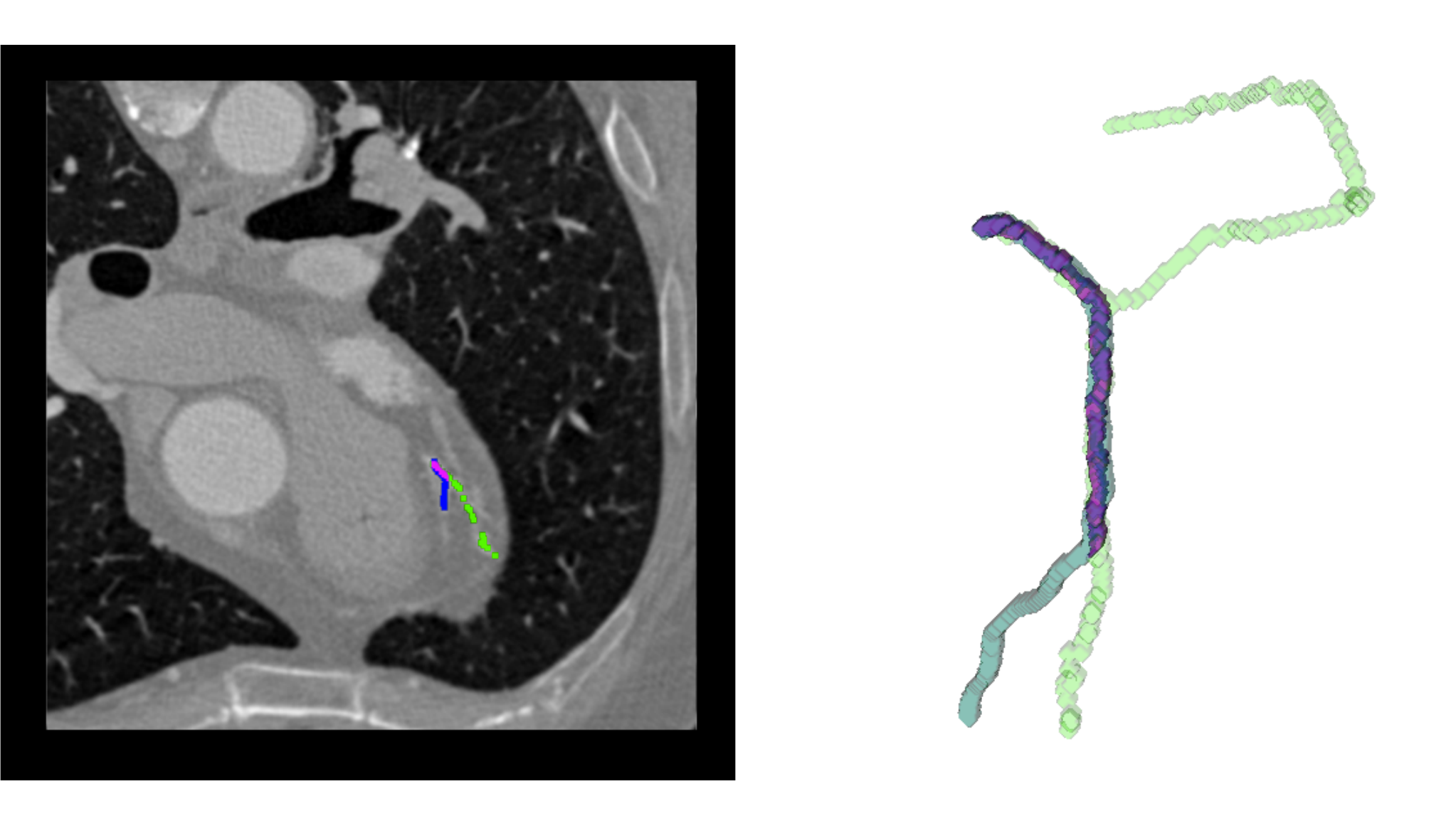}
\caption{This figure shows what we called the blind effect using an axial view and the 3D model generated. It is possible to visually understand that since our algorithm does not take into account a specific vessel end point to follow a vessel path, it will not necessarily find a desired vessel, but segment all the vessels connected to the given start point.}
\label{fig:blind}
\end{figure}

Considering the single vessel reference available for this dataset, an interesting effect is noticed. Since we use a single start point for segmenting the vascular network, a so-called "blind effect" is observed. Even though the segmentation process segments the network and includes the reference single vessel, it also segments other branches, and therefore would be badly evaluated by the competition evaluation tool. If we restrict the algorithm to find a single path, then it is not possible to ensure a-priori which of the three detected branches would be chosen. Figure \ref{fig:blind} depicts the effect.

\subsubsection{Carotids Data}

Carotids are the vessels that irrigate the brain, and therefore are very important for many medical conditions such as the cerebrovascular accidents, commonly known as strokes. We also tested our algorithm in segmenting the carotid vessels, using the dataset \cite{carotid} created for the 3rd MICCAI Workshop in the series "3D Segmentation in the Clinic: a Grand Challenge III". 

\begin{figure}[t]
\centering
\includegraphics[scale=0.5]{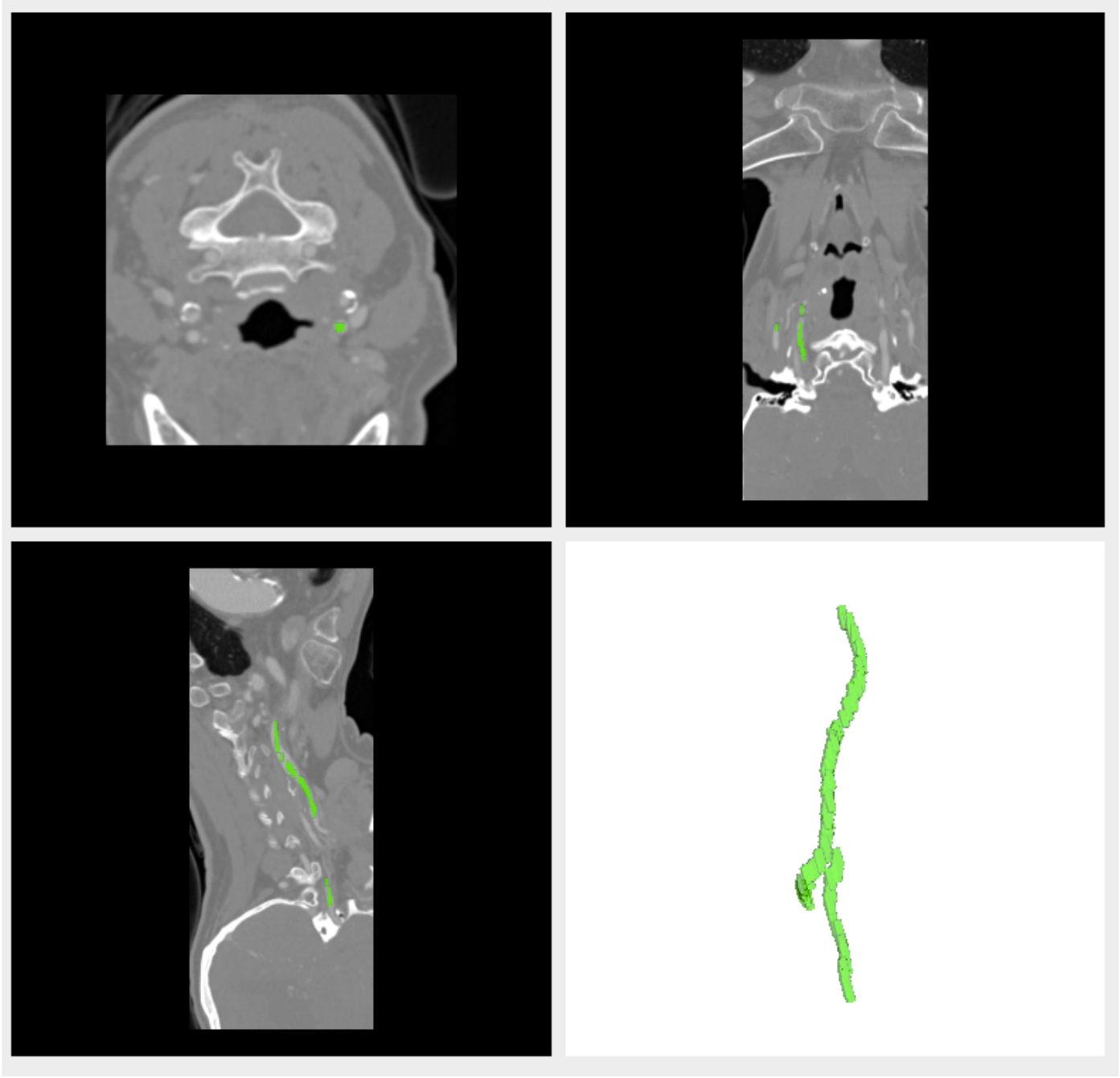}
\caption{Vascular segmentation for carothids real dataset. We see the results in the axial, coronal and sagittal views and the 3D model generated.}
\label{fig:carothid}
\end{figure}

The reference data available for this challenge concerns the identification of the carotid bifurcation. Even though the carotid vascular network reference is not available, it is possible to assess visually the outcome, since these vessels are morphologically simple, usually having a single main bifurcation. Figure \ref{fig:carothid} shows that both vessels and the bifurcation are found.

\subsubsection{Olfactory Projection Fibers (OPF)}

The OPF dataset is actually not from a vascular system, but from a nervous fibers network of the olfactory system. This, of course, hinders the detection of the vascular network (the use of a more appropriate model for vesselness would be advisable), albeit good results were achieved. The dataset is available at the DIADEM (short for Digital Reconstruction of Axonal and Dendritic Morphology) challenge website \cite{diadem}, which was a competition for evaluating algorithmic methods for automated neuronal tracing. 

OPF are network-like structures, but the inner part, corresponding to the lumen in vessels, are not exactly homogeneous and therefore the Gaussian mixture model proposed for vesselness evaluation, delivers low likelihood values. Still, it is possible to take advantage of the structure and segment at least part of the network. 

We used this non-vascular network dataset in our experiments mainly because of its reference data, which allowed us to evaluate our method quantitatively. For this, we used the evaluation tool proposed for the DIADEM challenge.

\begin{figure}[t]
\centering
\includegraphics[scale=0.4]{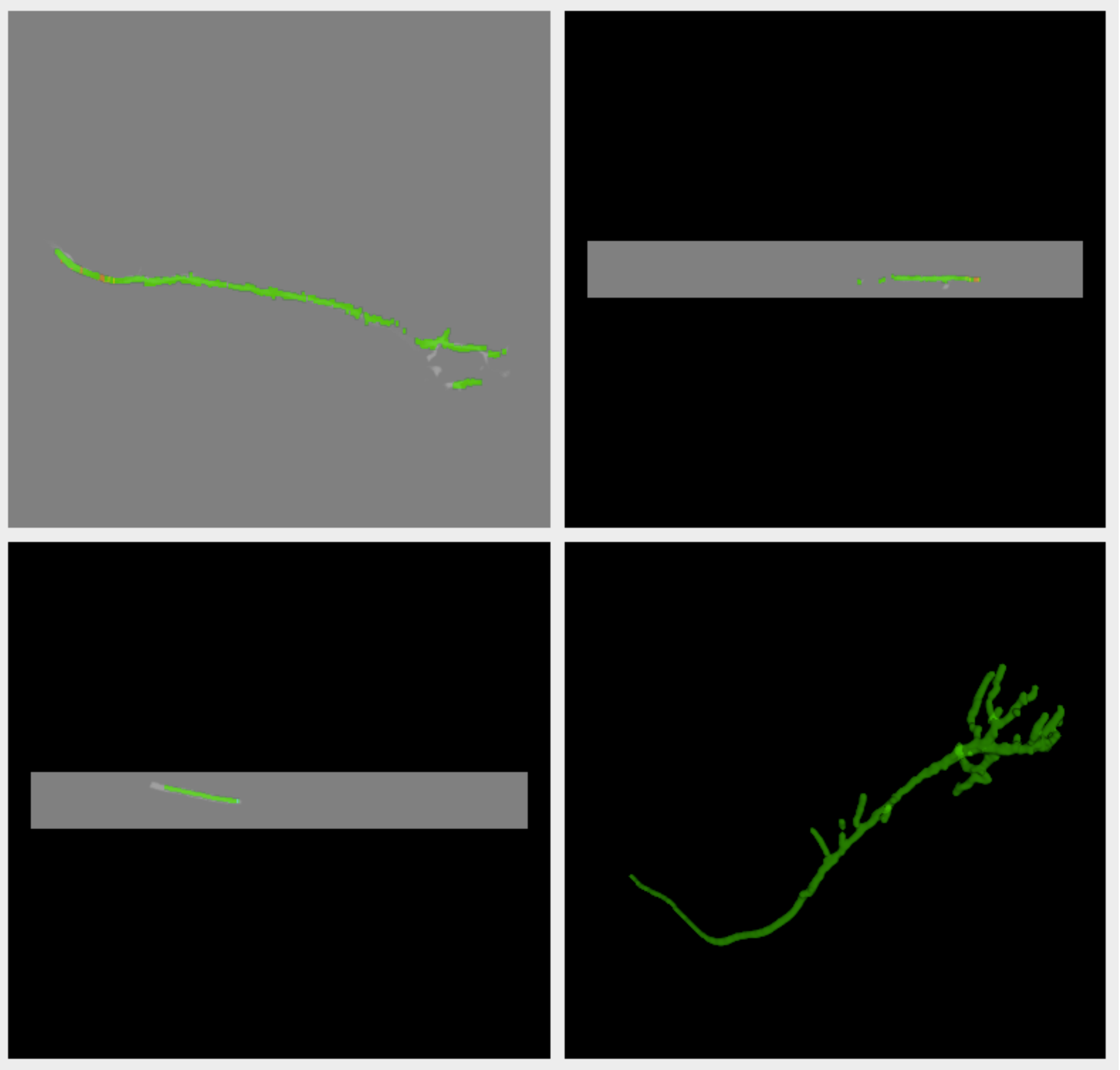}
\caption{Segmentation results for OPF dataset. We see the results in the axial, coronal and sagittal views and the 3D model generated.}
\label{fig:opf}
\end{figure}

The obtained results are summarized in Table \ref{tab:opftab}, which allows for a comparison of our method with other works. In the table we compare with \cite{turetkenneuro2011, turetkencvpr2012}, but the interested reader can refer to the challenge website to check other methods results. The results obtained are comparable with the ones obtained by methods designed for segmenting nervous fibers networks. This is encouraging and we believe that a more suited metric for nervous fibers (instead of the proposed vesselness) would improve substantially the results, even though this is not in the scope of this paper. Figure \ref{fig:opf} gives a visual feedback of the results obtained in one of the available datasets.

\begin{table}[!htp]
\small
\begin {center}
  \begin{tabular}{ p{2.62cm}  p{0.85cm}  p{0.85cm}  p{0.85cm} p{0.85cm}  p{0.85cm} p{0.85cm} p{0.85cm}  p{0.85cm}  }
     &  {Exam1} & {Exam3} & {Exam4}  & {Exam5} &  {Exam6} & {Exam7} & {Exam8}  & {Exam9}   \\ \hline
    k-MST Türetken11 &  \hspace{0.2cm}--  &  \hspace{0.2cm}-- &  0.865 &  \hspace{0.2cm}-- & 0.898 &  \hspace{0.2cm}-- & 0.722 &  \hspace{0.2cm}--\\ 
    HGD-QMIP Türetken12&   \hspace{0.2cm}--  &  \hspace{0.2cm}-- &  0.923 &  \hspace{0.2cm}-- & 0.911 &  \hspace{0.2cm}-- & 0.722 &  \hspace{0.2cm}--\\  \hline 
    Proposed method &  0.800  & 0.818 &  0.745  &  & 0.833 &  0.843 & 0.692 & 0.327 \\   
        \end{tabular}
  \end{center}
    \caption{OPF database results. The table shows the results obtained for each exam available in the website. The metric is the one made available for the competition, therefore we have a straight comparison with other methods.}
\label{tab:opftab}
\end{table}
\normalsize

\clearpage

\subsection{Topological Description}
\label{sec:resultado:topology}

Our method delivers not only the vascular network segmentation but its topology as well. It is possible to detect vascular branches and identify connecting points, which are assumed to be bifurcations (even though the exact point of a bifurcation is not very accurately determined). 

\begin{figure}[t]
\centering
\includegraphics[scale=0.35]{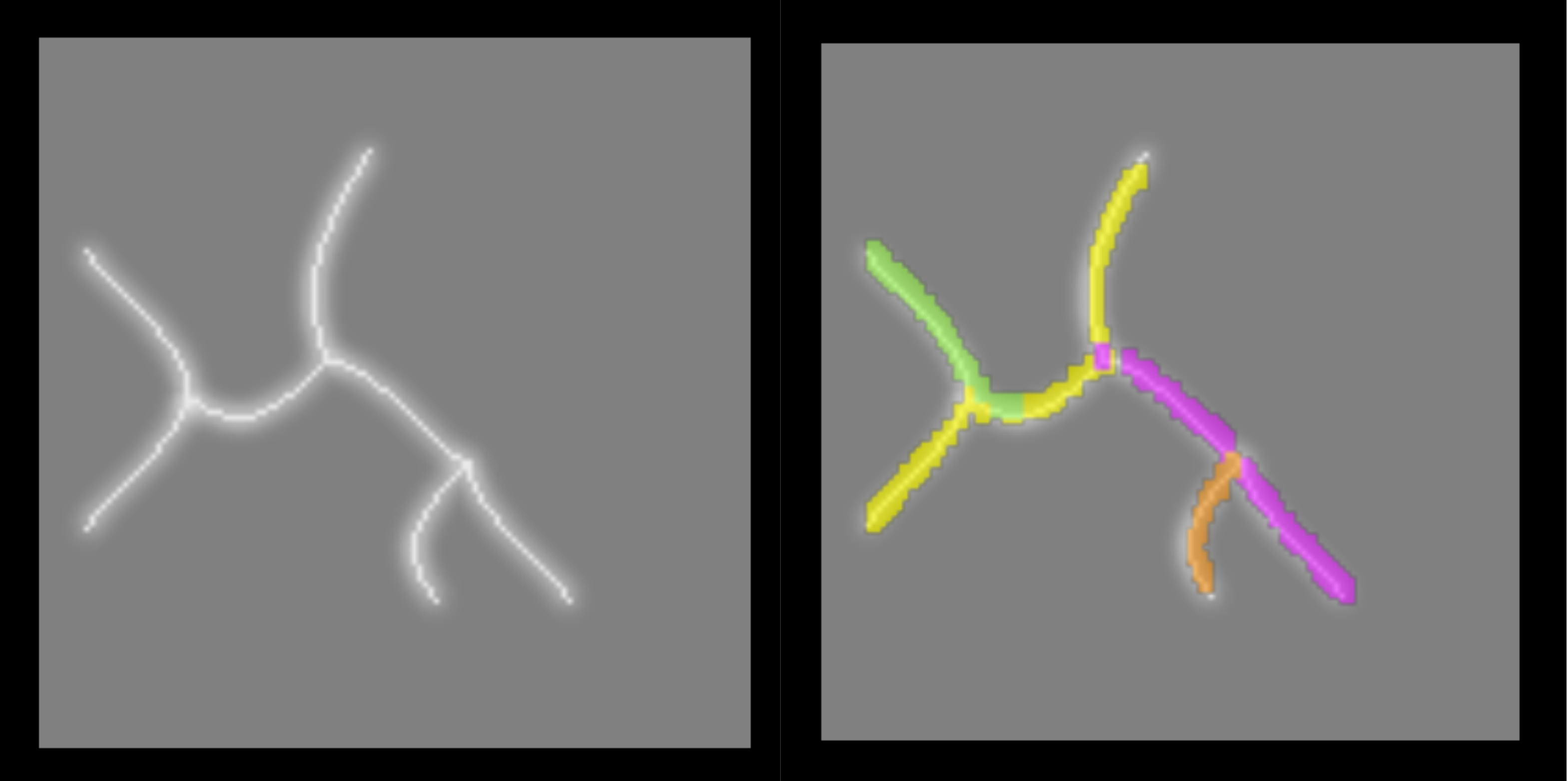}
\includegraphics[scale=0.35]{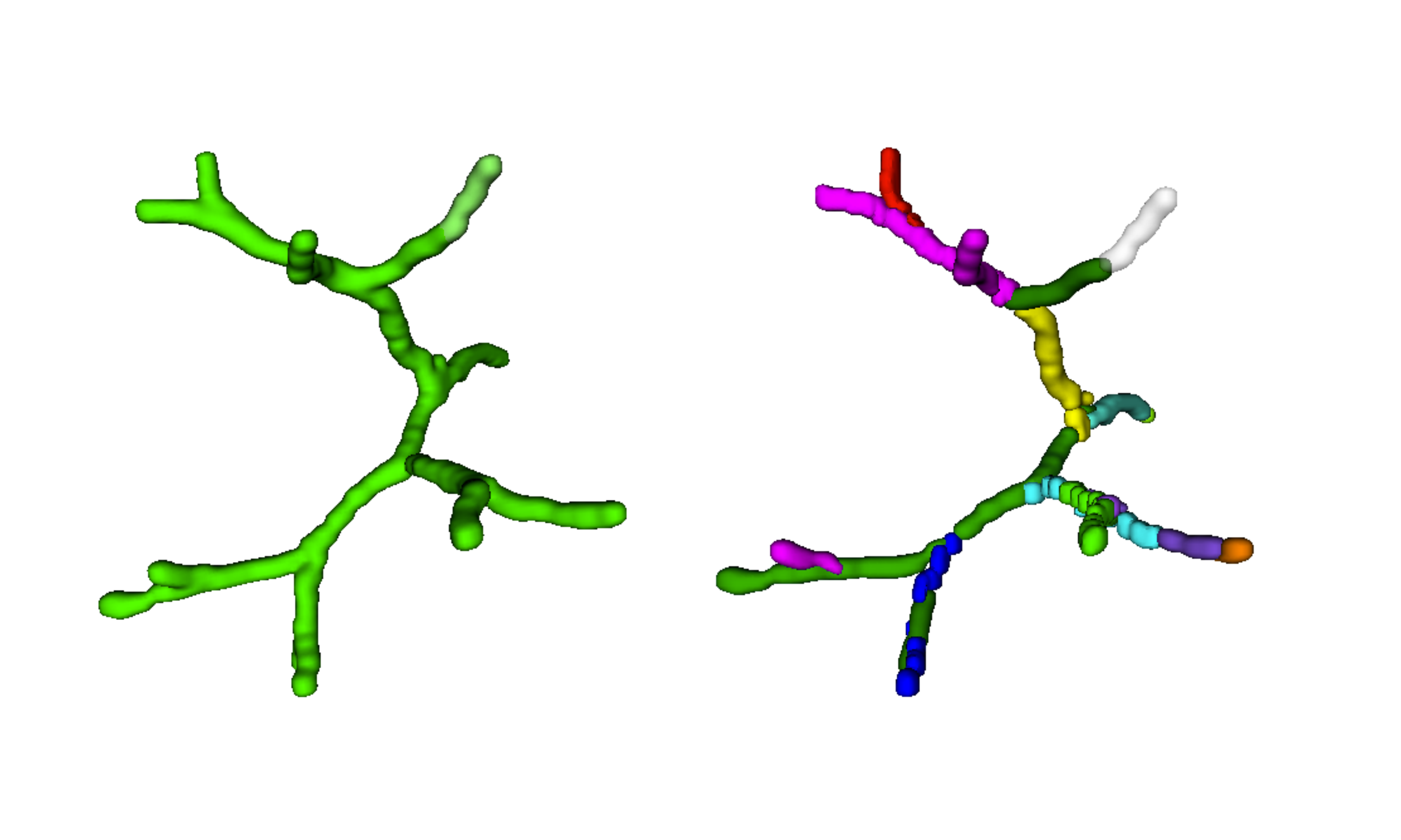}
\caption{Extracted topology for synthetic data, where each color represents a different branch. In each row a different synthetic model is showed, the first a 2D synthetic model, and the second a 3D model, as referred previously.}
\label{fig:topology1}
\end{figure}

We have not evaluated nummerically the results for topology with the datasets used, but some visual feedback is given in figure \ref{fig:topology1}. It is possible to visualize branches in different colours and therefore inspect visually the vascular network topology. The results are coherent with visual inspection.

\clearpage

\section{Conclusions}
\label{sec:conclusions}
This paper presents a method to segment full vascular networks through an iterative procedure that starts at a single vessel point provided by the user. A cloud of samples is defined within conical volume having at its apex the current vessel point. The decision whether or not a sample is a vessel point is based on a metric of how well the sample fits a vessel model. The vascular network is modeled by a directed graph. The final vascular network connecting the detected vessel points is obtained by solving a a flow problem using linear programming. This procedure finds branches of vessels iteratively until no new branch is found.

We tested our method using several datasets. An extensive performance comparison of our method with alternative approaches could not be done in the present work, due the lack of a public data set for full vascular network segmentation. All publicly available data sets just contain single vessel branches, or non vascular network structures, such as a nervous fiber network.

Synthetic data was correctly segmented even for datasets with many bifurcations and vessels with different diameters. The method also produced visually coherent vascular networks for different organs. A dataset of pulmonary CT images containing vascular networks with high capillarity, was reasonably segmented, specially if one considers that a single seed point was used. Datasets of carotid and coronary CT scans stressed the algorithm due to the very nature of their data with many different structures surrounding the vessels, some of them sharing vascular textures and densities. Nevertheless, the results obtained were visually coherent with an specialist expectancy. Finally, we tested our algorithm using an OPF dataset, which, despite not being a vascular dataset but a nervous fibers network, provided a reference with which we could evaluate numerically our method. Even though the vesselness measurement is not very suited for evaluating nervous fibers, good results were obtained and the olfactory fibers network was segmented properly. The results obtained show the potential of the proposed method to segment and extract the topology of different vascular networks. 


\afterpage{\clearpage}

\section*{Acknowledgements}
  The authors acknowledge Faperj, CNPq and CAPES for funding this research.



\bibliographystyle{model2-names}

\bibliography{sample}

\end{document}